\documentclass[10pt,journal,compsoc]{IEEEtran}

\ifCLASSOPTIONcompsoc
  \usepackage[nocompress]{cite}
\else
  \usepackage{cite}
\fi
\ifCLASSINFOpdf
\else
\fi

\usepackage{multirow}
\usepackage{amsmath}
\usepackage{amsthm}
\usepackage{stackengine}
\usepackage{amssymb}
\usepackage{algorithm}
\usepackage{algorithmicx}
\usepackage{algpseudocode} 
\usepackage{makecell}
\usepackage{color}
\usepackage{tikz}
\usepackage{caption}
\usepackage{lscape}
\usepackage{pdflscape}
\usepackage{rotating}
\usepackage{algpseudocode}
\usepackage{empheq} 
\usepackage{color,soul}
\usepackage{hyperref}

\usepackage{amsmath}

\usetikzlibrary{positioning, automata, chains, arrows.meta}
\usetikzlibrary{matrix,positioning,calc}
\newcommand{\True}{\mbox{{1}}}
\newcommand{\False}{\mbox{{0}}}

\newtheorem{mytheorem}{Theorem}

\newtheorem{remark}{Remark}

\renewcommand{\qedsymbol}{$\blacksquare$}

\algnewcommand{\LineComment}[1]{\State \(\triangleright\) #1}

\newcommand{\floor}[1]{\left\lfloor #1\right\rfloor}

\begin{document}
%
\title{On the Convergence of Tsetlin Machines for \\the AND and the OR Operators}
%
%
%
%

\author{Lei~Jiao,~\IEEEmembership{Senior Member,~IEEE}, ~Xuan~Zhang,~Ole-Christoffer Granmo
\IEEEcompsocitemizethanks{\IEEEcompsocthanksitem Lei Jiao and Ole-Christoffer Granmo are with the Centre for Artificial Intelligence Research, University of Agder, 4879, Grimstad, Norway.\\
E-mail: lei.jiao@uia.no; ole.granmo@uia.no
\IEEEcompsocthanksitem Xuan Zhang is with the Norwegian
Research Centre (NORCE), 4879, Grimstad, Norway. \\E-mail: xuzh@norceresearch.no}
}

\IEEEtitleabstractindextext{%
\begin{abstract}
The Tsetlin Machine (TM) is a novel machine learning algorithm based on propositional logic, which has obtained the state-of-the-art performance on several pattern recognition problems. In previous studies, the convergence properties of TM for 1-bit operation and XOR operation have been analyzed. To make the analyses on the basic digital operations complete, in this article, we analyze the convergence when input training samples follow AND and OR operators respectively. Our analyses reveal that the TM can converge almost surely to reproduce AND and OR operators, which are learnt from training data over an infinite time horizon. Specifically, by analysing the OR operator, we reveal the convergence property of TM when two sub-patterns can be jointly represented by one clause, which is quite distinct compared with the analysis of the XOR case.  
The analyses on AND and OR operators, together with the previously analysed 1-bit and XOR operations, complete the convergence analyses on basic operators in Boolean algebra. 
\end{abstract}

}

\maketitle

\IEEEdisplaynontitleabstractindextext

%
\IEEEpeerreviewmaketitle

\section{Introduction}


The Tsetlin Machine (TM)~\cite{granmo2018tsetlin} organizes groups of Tsetlin Automata (TAs)~\cite{Tsetlin1961} to collaboratively learn distinct patterns in training data. A TA, which is the core learning entity of TM, is a kind of learning automata that select their current actions based on past experiences learnt from the environment in order to obtain the maximum reward. The state of the art of the study in learning automate is presented in~\cite{zhang2019conclusive,yazidi2019hierarchical}. In TM,  each group of TAs builds a clause in propositional logic, which captures a specific sub-pattern. Specifically, a clause is a conjunction of literals, where a literal is a propositional input or its negation. Once distinct sub-patterns are learnt by a number of clauses, the overall pattern recognition task is completed by a voting scheme from the clauses.

TMs possess two main advantages: transparent inference and hardware friendliness~\cite{granmo2018tsetlin}. TMs provide transparent learning that shows how sub-patterns are composed in clauses, by including or excluding certain literals based on the feedback that they receive. Employing propositional logic for knowledge representation provides rules rather than a mathematical computation, which is advantageous compared with computation based approaches, such as deep learning. In fact, for the deep learning based models, such as attention mechanisms, the learning process itself remains inside a black-box neural network. In general, attention weights do not provide a meaningful explanation~\cite{ribeiro2016should, jain2019attention}. Unlike neural networks, the propositional form of TMs, with non-monotone clauses, is human-interpretable~\cite{reiter1988nonmonotonic}. Computationally, TMs is composed by a set of finite-state automata that are in nature appropriate for hardware implementation~\cite{wheeldon2020learning}. Different from the comprehensive arithmetic operations required by most other AI algorithms, only increment and decrement operations are sufficient for a TA in TM to learn~\cite{Tsetlin1961}, which is indeed hardware friendly. 

There are many variations of TMs with two main architectures: the convolutional TM (CTM)~\cite{granmo2019convolutional} and the regression TM (RTM)~\cite{ abeyrathna2019nonlinear,abeyrathna2020integerregression}. The TM, together with its variations,  has been employed in several applications, such as word sense disambiguation~\cite{yadav2021interpretability}, aspect-based sentiment analysis~\cite{rohan2021AAAI}, novelty detection~\cite{icaart21bimal},  text classification~\cite{Rohanblackbox} with enhanced interpretability~\cite{yadav2022robustness}, and solving contextual bandit problems~\cite{RaihanNIPS22}. The CUDA version of TM~\cite{abeyrathna2020massively} is also developed, which makes the TM notably more applicable by speeding up TM learning through a novel parallelization scheme, where each clause runs and learns independently in its own thread.  The above studies indicate that TMs obtain better or competitive classification and regression accuracy compared with most of the state-of-the-art techniques. At the same time, the transparency of learning is maintained with smaller memory footprint and higher computational efficiency. 

For the convergence analysis of TM, the convergence properties of the 1-bit operator and the XOR operator have been analyzed in~\cite{zhang2020convergence} and~\cite{jiao2021convergence} respectively. In more details, the convergence for unary operators on one-bit data, i.e., the IDENTITY- and the NOT operators, is analyzed in~\cite{zhang2020convergence}, where we first prove that the TM can converge almost surely to the intended pattern when the training data is noise-free. Thereafter, we analyze the effect of noise, establishing how the noise in the data and the granularity parameter of the TM, $s$, govern convergence~\cite{zhang2020convergence}. Through the study of the 1-bit case, the functionalities of the hyper-parameter $s$ for granularity of learning and the chain length of the TA are thoroughly revealed. The convergence property of XOR operator is studied in~\cite{jiao2021convergence}. For XOR operator, there are two sub-patterns, namely (0, 1) and (1, 0), with non-linear relationship. We first prove the convergence of a simple structure with two clauses, each of which has four TAs with two states. Markov chain analysis is adopted for that proof. The analysis indicates that even that simple structure can guarantee the TM to learn the intended XOR logic almost surely. Thereafter, we study the convergence behavior of a more general case, where multiple clauses exist. Through the latter analysis, we reveal the crucial role the hyper-parameter $T$ plays in TM, showing how this hyper-parameter balances the clauses to robustly capture distinct sub-patterns within one class.

\textbf{Paper Contributions.} Based on the summary of the existing analytical work of TM, we understand that 
the convergence study of the basic logic operators is not complete. In this paper, we will fill in the gap and analyze the convergence of TM for AND and OR operators.
This paper, together with~\cite{zhang2020convergence} and~\cite{jiao2021convergence}, completes the study on the convergence analysis of all fundamental Boolean operators in TM, offering a conclusive convergence analysis and establishing a milestone of TM studies.   Additionally, because the OR operator offers the possibility
to jointly represent two sub-patterns by one clause, the convergence analysis of the OR operator is quite different compared with the XOR case, which adds new insights to the convergence nature of the TM.

\textbf{Paper Organization.} The remaining of the paper is organized as follows. Section \ref{sect:TM} briefly reviews the TM and specifies the training process. In Section~\ref{sec:AND}, we present our analytical procedure and the main analytical results for the AND operator. Section~\ref{Sec:OR} analyzes the convergence for the OR operator before we conclude the paper in Section \ref{conclusions}.

\section{Brief Overview of the TM}\label{sect:TM}
To make the article self-contained, we present the basics of TM here in this section. Note that this section is technically identical to Section 2 of article~\cite{jiao2021convergence}. Those who already are familiar with the concept and notations of TM can jump directly to Section~\ref{sec:AND}. 

\subsection{Review of Tsetlin Machines}

A TM that is to learn the characteristics of class $i$ is formed by $m$ teams of TAs. More specifically, the TM trains TAs to formulate $m$ clauses (TA teams), $C^i_j,~j=1,2,...,m$, and to capture the sub-patterns that characterize the class $i$. The input of a TM is denoted by $\bold{X}=[x_1, x_2, \ldots, x_o]$, $x_k \in \{0, 1\}$, $k=1, 2, \ldots, o$. Each TA team contains $o$ pairs of TAs, and thus a TA team $\mathcal{G}^i_j=\{\mathrm{TA}^{i,j}_{k'}|1\leq k'\leq 2o\}$ has $2o$ TAs. 
 For a certain input variable $x_k$, there is a pair of TAs that are responsible for its role in the clause. The automaton $\mathrm{TA}^{i,j}_{2k-1}$ is responsible for the original form of the input $x_k$, whereas $\mathrm{TA}^{i,j}_{2k}$ addresses the negation of $x_k$, i.e., $\neg x_k$. Note that the original forms of the inputs and their negations are jointly referred to as literals.

\begin{figure*}
\centering
\resizebox{1\textwidth}{!}{
\begin{minipage}{1\textwidth}
\begin{tikzpicture}[node distance = .35cm, font=\Huge]
    \tikzstyle{every node}=[scale=0.35]
    \node[state] (A) at (0,2) {~~~~0~~~~~};
    \node[state] (B) at (1.5,2) {~~~~1~~~~~};
    
    \node[state,draw=white] (M) at (3,2) {~~~$....$~~~};
    
    \node[state] (C) at (4.5,2) {$~N-2~$};
    \node[state] (D) at (6,2) {$~N-1~$};
    
    \node[state] (E) at (7.5,2) {$~~~~N~~~~$};
    \node[state] (F) at (9,2) {$~N+1~$};
    
    \node[state,draw=white] (G) at (10.5,2) {~~~$....$~~~};
    
    \node[state] (H) at (12,2) {$2N-2$};
    \node[state] (I) at (13.5,2) {$2N-1$};

    \node[thick] at (4,4) {$Action~1$};
    \node[thick] at (9.5,4) {$Action~2$};
    
    \node[thick] at (1.9,1) {$Reward~(R):~\dashrightarrow$~~~~$Penalty~(P):~\rightarrow$};

    \draw[every loop]
    (A) edge[bend left] node [scale=1.2, above=0.1 of B]{} (B)
    (B) edge[bend left] node  [scale=1.2, above=0.1 of M] {} (M)
    (M) edge[bend left] node  [scale=1.2, above=0.1 of C] {} (C)
    (C) edge[bend left] node [scale=1.2, above=0.1 of D] {} (D)
    (D) edge[bend left] node  [scale=1.2, above=0.1 of E] {} (E);

    \draw[every loop]
    (I) edge[bend left] node [scale=1.2, below=0.1 of H] {} (H)
    (H) edge[bend left] node  [scale=1.2, below=0.1 of G] {} (G)
    (G) edge[bend left] node [scale=1.2, below=0.1 of F] {} (F)
    (F) edge[bend left] node  [scale=1.2, below=0.1 of E] {} (E)
    (E) edge[bend left] node  [scale=1.2, below=0.1 of D] {} (D);

    
    \draw[dashed,->]
    (B) edge[bend left] node  [scale=1.2, above=0.1 of A] {} (A)
    (M) edge[bend left] node [scale=1.2, above=0.1 of B] {} (B)
    (C) edge[bend left] node [scale=1.2, above=0.1 of M] {} (M)
    (D) edge[bend left] node [scale=1.2, above=0.1 of C] {} (C)
    (A) edge[loop left] node [scale=1.2, below=0.1 of D] {} (D);
    
    \draw[dashed,->]

    (H) edge[bend left ] node [scale=1.2, below=0.1 of I] {} (I)
    (G) edge[bend left] node  [scale=1.2, below=0.1 of H] {} (H)
    (F) edge[bend left] node  [scale=1.2, below=0.1 of G] {} (G)
    (E) edge[bend left ] node [scale=1.2, below=0.1 of F] {} (F)
    (I) edge[loop right] node [scale=1.2, below=0.1 of E] {} (E);
    
      \draw[dotted, thick] (6.75,0.6) -- (6.75,3);

\end{tikzpicture}
\end{minipage}
}
\caption{A two-action Tsetlin automaton with $2N$ states~\cite{jiao2021convergence}.}
\label{figure:TAarchitecture_basic}
\end{figure*}
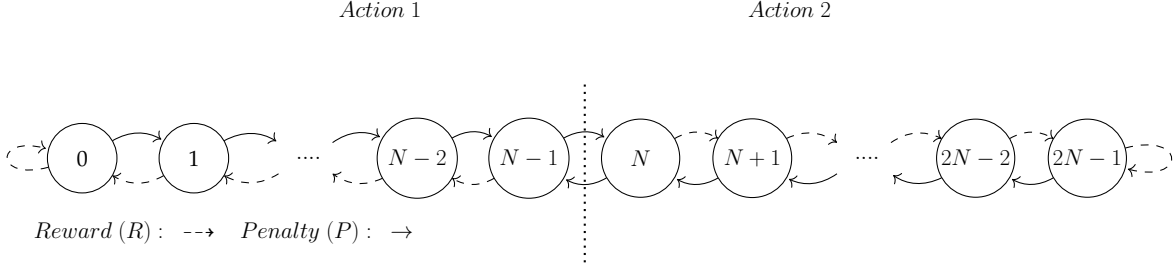

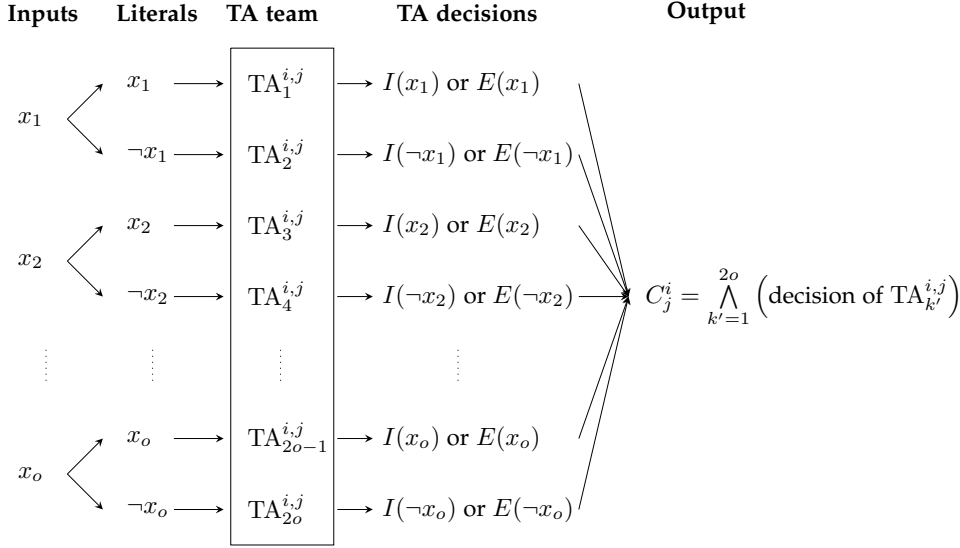
\begin{figure*}[htbp]
\begin{centering}
\resizebox{1.37\textwidth}{!}{
\begin{minipage}{1\textwidth}
\resizebox{0.53\textwidth}{!}{
\begin{tikzpicture}[node distance = .35cm]

    \node[label=left:{\bf Inputs}] at (0.2,1.45) {};
    \node[label=right:{\bf Literals}] at (0.25,1.5) {};
    \node[label=right:{\bf TA team}] at (1.8,1.5) {};
    \node[label=right:{\bf TA decisions}] at (4.2,1.5) {};
    \node[label=right:{\bf Output}] at (8.,1.5) {};

    \node[label=left:$x_1$] at (-0.3,0) {};
    \node[label=left:$x_2$] at (-0.3,-2) {};
    \node[label=left:$x_o$] at (-0.3,-5) {};
    
    \node[label=right:$x_1$] at (0.4,0.5) {};
    \node[label=right:$\neg x_1$] at (0.4,-0.5) {};
    \node[label=right:$x_2$] at (0.4,-1.5) {};
    \node[label=right:$\neg x_2$] at (0.4,-2.5) {};
    \node[label=right:$x_o$] at (0.4,-4.5) {};
    \node[label=right:$\neg x_o$] at (0.4,-5.5) {};
    
    \node[label=right:$\mathrm{TA}_1^{i,j}$] at (2.1,0.5) {};
    \node[label=right:$\mathrm{TA}_2^{i,j}$] at (2.1,-0.5) {};
    \node[label=right:$\mathrm{TA}_3^{i,j}$] at (2.1,-1.5) {};
    \node[label=right:$\mathrm{TA}_4^{i,j}$] at (2.1,-2.5) {};
    \node[label=right:$\mathrm{TA}_{2o-1}^{i,j}$] at (2.1,-4.5) {};
    \node[label=right:$\mathrm{TA}_{2o}^{i,j}$] at (2.1,-5.5) {};
    \draw (2.1, 1) -- (3.55, 1) -- (3.55, -6) -- (2.1, -6) -- (2.1, 1); 
    
    \node[label=right:$I(x_1)~\text{or}~ E(x_1)$] at (4.,0.5) {};
    \node[label=right:$I(\neg x_1)~\text{or}~ E(\neg x_1)$] at (4.,-0.5) {};
    \node[label=right:$I(x_2)~\text{or}~ E(x_2)$] at (4.,-1.5) {};
    \node[label=right:$I(\neg x_2)~\text{or}~ E(\neg x_2)$] at (4.,-2.5) {};
    \node[label=right:$I(x_o)~\text{or}~ E(x_o)$] at (4.,-4.5) {};
    \node[label=right:$I(\neg x_o)~\text{or}~ E(\neg x_o)$] at (4.,-5.5) {};
    
    \node[label=right:{$C^i_j=\bigwedge\limits_{k'=1}^{2o} \left(\text{decision of}~ \mathrm{TA}_{k'}^{i,j}\right)$}] at (7.7,-2.5) {};
    
    \draw [-{stealth[length=4mm]}] (-0.2,0) -- (0.3,0.5);
    \draw [-{stealth[length=4mm]}] (-0.2,0) -- (0.3,-0.5);
    \draw [-{stealth[length=4mm]}] (-0.2,-2) -- (0.3,-1.5);
    \draw [-{stealth[length=4mm]}] (-0.2,-2) -- (0.3,-2.5);
    \draw [-{stealth[length=4mm]}] (-0.2,-5) -- (0.3,-4.5);
    \draw [-{stealth[length=4mm]}] (-0.2,-5) -- (0.3,-5.5);
    
    \draw [-{stealth[length=4mm]}] (1.3,0.5) -- (2., 0.5);
    \draw [-{stealth[length=4mm]}] (1.3,-0.5) -- (2., -0.5);
    \draw [-{stealth[length=4mm]}] (1.3,-1.5) -- (2, -1.5);
    \draw [-{stealth[length=4mm]}] (1.3,-2.5) -- (2., -2.5);
    \draw [-{stealth[length=4mm]}] (1.3,-4.5) -- (2., -4.5);
    \draw [-{stealth[length=4mm]}] (1.3,-5.5) -- (2., -5.5);
    
    \draw [-{stealth[length=4mm]}] (3.6,0.5) -- (4.1, 0.5);
    \draw [-{stealth[length=4mm]}] (3.6,-0.5) -- (4.1, -0.5);
    \draw [-{stealth[length=4mm]}] (3.6,-1.5) -- (4.1, -1.5);
    \draw [-{stealth[length=4mm]}] (3.6,-2.5) -- (4.1, -2.5);
    \draw [-{stealth[length=4mm]}] (3.6,-4.5) -- (4.1, -4.5);
    \draw [-{stealth[length=4mm]}] (3.6,-5.5) -- (4.1, -5.5);
    
    \draw [-{stealth[length=3mm]}] (7.,0.5) -- (7.7, -2.5);
    \draw [-{stealth[length=3mm]}] (7.,-0.5) -- (7.7, -2.5);
    \draw [-{stealth[length=3mm]}] (7.,-1.5) -- (7.7, -2.5);
    \draw [-{stealth[length=3mm]}] (7.,-2.5) -- (7.7, -2.5);
    \draw [-{stealth[length=3mm]}] (7.,-4.5) -- (7.7, -2.5);
    \draw [-{stealth[length=3mm]}] (7.,-5.5) -- (7.7, -2.5);

    \draw
    [dotted] (-0.5,-3.25) -- (-0.5,-3.75)
    [dotted] (1.,-3.25) -- (1.,-3.75)
    [dotted] (2.8,-3.25) -- (2.8,-3.75)
    [dotted] (5.3,-3.25) -- (5.3,-3.75);
    
\end{tikzpicture}
}

\end{minipage}
}
\end{centering}
\caption{\label{fig:tateam} A TA team $G^i_j$ consisting of $2o$ TAs \cite{zhang2020convergence}. Here $I(x_1)$ means ``include $x_1$'' and $E(x_1)$ means ``exclude $x_1$''.  }
\end{figure*}

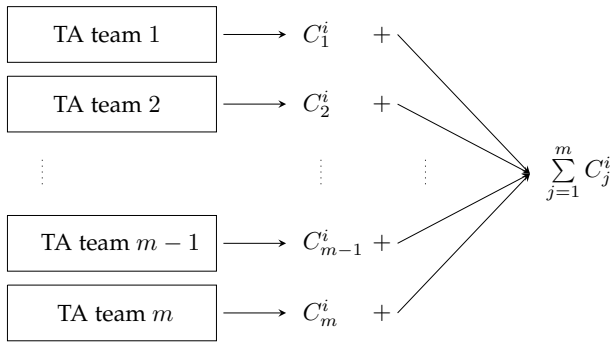
\begin{figure}[htbp]
\begin{center}
\resizebox{0.38\textwidth}{!}{
\begin{minipage}{0.4\textwidth}
\begin{tikzpicture}[node distance = .35cm]

\node[label=left:TA team $1~~~~~~$] at (0,0) {};
\node[label=left:TA team $2~~~~~~$] at (0,-1) {};
\node[label=left:TA team $m-1$] at (0,-3) {};
\node[label=left:TA team $m~~~~$] at (0,-4) {};
\draw (-3, 0.4) -- (-3, -0.4) -- (0, -0.4) -- (0, 0.4) -- (-3, 0.4);
\draw (-3, -0.6) -- (-3, -1.4) -- (0, -1.4) -- (0, -0.6) -- (-3, -0.6);
\draw (-3, -2.6) -- (-3, -3.4) -- (0, -3.4) -- (0, -2.6) -- (-3, -2.6);
\draw (-3, -3.6) -- (-3, -4.4) -- (0, -4.4) -- (0, -3.6) -- (-3, -3.6);

\node[label=right: $C^i_1$] at (1,0) {};
\node[label=right: $C^i_2$] at (1,-1) {};
\node[label=right: $C^i_{m-1}$] at (1,-3) {};
\node[label=right: $C^i_m$] at (1,-4) {};
\node[label=right:$+$] at (2,0) {};
\node[label=right:$+$] at (2,-1) {};
\node[label=right:$+$] at (2,-3) {};
\node[label=right:$+$] at (2,-4) {};

\node[label=right: $\sum\limits_{j=1}^{m} C^i_j$] at (4.5,-2) {};
    
\draw [-{stealth[length=4mm]}] (0.1,0) -- (1,0);
\draw [-{stealth[length=4mm]}] (0.1,-1) -- (1,-1);
\draw [-{stealth[length=4mm]}] (0.1,-3) -- (1,-3);
\draw [-{stealth[length=4mm]}] (0.1,-4) -- (1,-4);

\draw [-{stealth[length=4mm]}] (2.6,0) -- (4.5, -2);
\draw [-{stealth[length=4mm]}] (2.6,-1) -- (4.5, -2);
\draw [-{stealth[length=4mm]}] (2.6,-3) -- (4.5, -2);
\draw [-{stealth[length=4mm]}] (2.6,-4) -- (4.5, -2);
    
\draw [dotted] (-2.5,-1.8) -- (-2.5,-2.2);
\draw [dotted] (1.5,-1.8) -- (1.5,-2.2);
\draw [dotted] (3,-1.8) -- (3,-2.2);

\end{tikzpicture}

\end{minipage}
}
\end{center}
\caption{\label{fig:TMVoting} TM voting architecture~\cite{jiao2021convergence}.}
\end{figure}

{\color{black}Each TA chooses one of two actions, i.e., it either ``Includes'' or ``Excludes'' its literal.  Figure \ref{figure:TAarchitecture_basic} illustrates the structure of a TA with two actions. When the TA is in any state on the left-hand side, i.e., $0$ to $N-1$, the action ``Include" is selected. The action becomes ``Exclude" when the TA is in a state on the right-hand side. The transitions among the states are triggered by a reward or a penalty that the TA receives from the environment, which, in this case, is determined by different types of feedback defined in the TM (to be explained later).     
Collectively, the outputs of the TA team take part in a conjunction, expressed by the conjunctive clause~\cite{zhang2020convergence}:
\begin{equation}
\label{eqn:clause1}
C^i_j(\bold{X}) = \begin{cases}
\left(\bigwedge\limits_{k \in I^i_j} {x_k}\right) \wedge \left(\bigwedge\limits_{k \in \bar{I}^i_j} {\neg x_k}\right) \wedge 1 & \mathrm{for\ training},\\
\left(\left(\bigwedge\limits_{k \in I^i_j} {x_k}\right) \wedge \left(\bigwedge\limits_{k \in \bar{I}^i_j} {\neg x_k}\right)\right) \vee 0 & \mathrm{for\ testing}.
\end{cases}
\end{equation}
In Eq. (\ref{eqn:clause1}), $I^i_j$ and $\bar{I}^i_j$ are the subsets of indexes for the literals that have been included in the clause. $I^i_j$ contains the indexes of included original (non-negated) inputs, $x_k$, whereas $\bar{I}^i_j$ contains the indexes of included negated inputs, $\neg x_k$. The ``0'' and ``1'' in Eq. (\ref{eqn:clause1}) make sure that $C^i_j(\bold{X})$ also is defined when all the TAs choose to exclude their literals. As can be observed, during training, an ``empty'' clause outputs $1$, while it outputs $0$ during testing and operation.
}

Figure \ref{fig:tateam} illustrates the structure of a clause and its relationship to its literals. Here, for  ease  of  notation in the analysis of the training procedure, let $I(x)=x, ~I(\neg x)=\neg x$, and $E( x)=E(\neg x)=1$, with the latter meaning that an excluded literal does not contribute to the output.


Multiple clauses, i.e., the TA teams in conjunctive form, are assembled into a complete TM. There are two architectures for clause assembling: Disjunctive Normal Form Architecture and Voting Architecture. In this study, we focus on the latter one, 
as shown in Figure~\ref{fig:TMVoting}. The voting consists of summing the outputs of the clauses:
\begin{equation}
\label{eqn:summation}
f_{\sum}(\mathcal{C}^i(\bold{X}))= \sum^m\limits_{j =1} C_j^i(\bold{X}).
\end{equation}
The output of the TM, in turn, is decided by the unit step function:
\begin{align}
\label{eqn:yivoting}
\hat{y}^i={\begin{cases}\False&{\text{for }}f_{\sum}(\mathcal{C}^i(\bold{X}))<Th\\\True&{\text{for }}f_{\sum}(\mathcal{C}^i(\bold{X}))\geq Th\end{cases}},
\end{align} 
where $Th$ is a predefined threshold for classification. Note that for voting architecture, the TM can assign a polarity to each TA team~\cite{granmo2018tsetlin}. For example, TA teams with odd indexes possess positive polarity, and they vote for class $i$. The remaining TA teams have negative polarity and vote against class $i$. The voting consists of summing the output of the clauses, according to polarity, and the threshold $Th$ is configured as zero. In this study, for ease of analysis, we consider only positive polarity clauses. Nevertheless, this does not change the nature of TM learning.


\subsection{The Tsetlin Machine for Learning Patterns}\label{sect:TMTraining}
\subsubsection{The Tsetlin Machine Game}\label{sect:TMGame}
The training process is built on letting all the TAs take part in a decentralized game. 
Training data $(\bold{X}=[x_1,x_2,...,x_o],~y^i)$ is obtained from a data set $\mathcal{S}$, distributed according to the probability distribution $P(\bold{X}, y^i)$.  In the game, each TA is guided by Type I Feedback and Type~II Feedback defined in Table \ref{table:type_i_feedback} and Table \ref{table:type_ii_feedback}, respectively. Type~I Feedback is triggered when the training sample has a positive label, i.e., $y^i=1$, meaning that the sample belongs to class $i$. When the training sample is labeled as not belonging to class $i$, i.e., $y^i=0$, Type II Feedback is utilized for generating responses. The parameter, $s$, controls the granularity of the clauses and a larger $s$ encourages more literals to be included in each clause. A more detailed analysis on parameter $s$ can be found in~\cite{zhang2020convergence}.



\begin{table}[h!]
\centering
\begin{tabular}{c|ccccc}
\multicolumn{2}{r|}{{\it Value of the clause} $C^i_j(\bold{X})$ }&\multicolumn{2}{c}{\True}&\multicolumn{2}{c}{\False}\\ 
\multicolumn{2}{r|}{{\it Value of the Literal} $x_k$/$\lnot x_k$}&{\True}&{\False}&{\True}&{\False}\\
\hline
\hline
\multirow{3}{*}{\bf Include Literal}&\multicolumn{1}{c|}{$P(\mathrm{Reward})$}&$\frac{s-1}{s}$&NA&$0$&$0$\\
&\multicolumn{1}{c|}{$P(\mathrm{Inaction})$}&$\frac{1}{s}$&NA&$\frac{s-1}{s}$&$\frac{s-1}{s}$\\
&\multicolumn{1}{c|}{$P(\mathrm{Penalty})$}&$0$&NA&$\frac{1}{s} $&$\frac{1}{s}$\\
\hline
\multirow{3}{*}{\bf Exclude Literal }&\multicolumn{1}{c|}{$P(\mathrm{Reward})$}&$0$&$\frac{1}{s}$&$\frac{1}{s}$ &$\frac{1}{s}$\\
&\multicolumn{1}{c|}{$P(\mathrm{Inaction})$}&$\frac{1}{s}$&$\frac{s-1}{s}$&$\frac{s-1}{s}$ &$\frac{s-1}{s}$\\
&\multicolumn{1}{c|}{$P(\mathrm{Penalty})$}&$\frac{s-1}{s}$&$0$&$0$&$0$\\
\hline
\end{tabular}
\caption{Type I Feedback --- Feedback upon receiving a sample with label $y=1$, for a single TA to decide whether to Include or Exclude a given literal $x_k/\neg x_k$ into $C^i_j$. NA means not applicable~\cite{granmo2018tsetlin}.}
\label{table:type_i_feedback}
\end{table}

\begin{table}[h!]
\centering
\begin{tabular}{c|ccccc}
\multicolumn{2}{r|}{\it Value of the clause $C^i_j(\bold{X})$}&\multicolumn{2}{c}{\True}&\multicolumn{2}{c}{\False}\\ 
\multicolumn{2}{r|}{\it Value of the Literal $x_k/\neg x_k$}&{\True}&{\False}&{\True}&{\False}\\
\hline
\hline
\multirow{3}{*}{\bf Include Literal }&\multicolumn{1}{c|}{$P(\mathrm{Reward})$}&$0$&$\mathrm{NA}$&$0$&$0$\\
&\multicolumn{1}{c|}{$P(\mathrm{Inaction})$}&$1.0$&$\mathrm{NA}$&$1.0$&$1.0$\\
&\multicolumn{1}{c|}{$P(\mathrm{Penalty})$}&$0$&$\mathrm{NA}$&$0$&$0$\\
\hline
\multirow{3}{*}{\bf Exclude Literal }&\multicolumn{1}{c|}{$P(\mathrm{Reward})$}&$0$&$0$&$0$&$0$\\
&\multicolumn{1}{c|}{$P(\mathrm{Inaction})$}&$1.0$&$0$&$1.0$ &$1.0$\\
&\multicolumn{1}{c|}{$P(\mathrm{Penalty})$}&$0$&$1.0$&$0$&$0$\\
\hline
\end{tabular}
\caption{Type II Feedback --- Feedback upon receiving a sample with label $y=0$, for a single TA to decide whether to Include or Exclude a given literal $x_k/\neg x_k$ into $C^i_j$. NA
means not applicable~\cite{granmo2018tsetlin}.}
\label{table:type_ii_feedback}
\end{table}

To avoid the situation that a majority of the TA teams learn only a small subset of the sub-patterns in the training data, forming an incomplete representation, we use a parameter $T$ as target for the summation $f_{\sum}$. If the votes for a certain sub-pattern already reach a total of $T$ or more, neither rewards nor penalties are provided to the TAs when more training samples of this particular sub-pattern are given. In this way, we can ensure that each specific sub-pattern can be captured by a limited number, i.e., $T$, of available clauses, allowing sparse sub-pattern representations among competing sub-patterns. In more details, the strategy works as follows:  


{\bf Generating Type I Feedback.} If the output from the training sample $\bold{X}$ is $y^i=\True$, we generate, in probability, \emph{Type I Feedback} for each clause $C^i_j \in \mathcal{C}^i$, where $\mathcal{C}^i$ is the set of clauses that are trained for class $i$. The probability of generating Type I Feedback is~\cite{granmo2018tsetlin}:
\begin{equation}
u_1=\frac{T - \mathrm{max}(-T, \mathrm{min}(T, f_{\sum}(\mathcal{C}^i(\bold{X}))))}{2T}.\label{u1}
\end{equation}



{\bf Generating Type II Feedback.} 
If the output of the training sample $\bold{X}$ is $y^i = \False$, we generate, again, in probability, \emph{Type II Feedback} to each clause $C^i_j \in \mathcal{C}^i$. The probability is~\cite{granmo2018tsetlin}:
\begin{equation}
u_2=\frac{T + \mathrm{max}(-T, \mathrm{min}(T, f_{\sum}(\mathcal{C}^i(\bold{X}))))}{2T}.\label{u2}
\end{equation}


After Type I Feedback or Type II Feedback is generated for a clause, the individual TA within each clause is given reward/penalty/inaction according to the probability defined, and then the states of the corresponding TAs are updated. 


\section{Convergence Analysis of TM for the AND Operator}\label{sec:AND}
In this section, we will prove the convergence of the AND operator. In this proof, we assume that the training samples do not have any noise, i.e.,
\begin{align}\label{andlogic}
P\left ( y=1 | x_{1}=1, x_{2}=1 \right ) = 1, \\
P\left ( y=0 | x_{1}=0, x_{2}=1 \right ) = 1,\nonumber\\
P\left ( y=0 | x_{1}=1, x_{2}=0 \right ) = 1,\nonumber\\
P\left ( y=0 | x_{1}=0, x_{2}=0 \right ) = 1.\nonumber 
\end{align}
We also assume that in the training samples, the above four cases will appear with non-zero probability. This means that all of the four types of training samples will appear during the training process. 

 Because the considered AND operator has only one pattern of input $\bold{X}$ that will trigger true output, we employ just one clause in this TM.   Because the considered TM has only one clause, for ease of expression, we ignore the indices of the classes and the clauses in our notation in the remainder of the proof. After simplification, $\mathrm{TA}^{i,j}_{k}$ becomes $\mathrm{TA}_{k}$, and $C_1^1$ becomes $C$. Since there are two input parameters, we configure 4 TAs, namely,  $\mathrm{TA}_1$, $\mathrm{TA}_2$, $\mathrm{TA}_3$, and $\mathrm{TA}_4$.  $\mathrm{TA}_1$ has two actions, i.e., including or excluding $x_{1}$. Similarly,  $\mathrm{TA}_2$ corresponds to including or excluding $\neg x_{1}$.  $\mathrm{TA}_3$ and $\mathrm{TA}_4$ determine the behavior of the $x_2$ and $\neg x_2$, respectively.   
 
 Clearly, there is only one pattern that will trigger the output $y$ to be $1$,  i.e., $x_{1}=1, x_{2}=1$, which corresponds to the logic to both include $x_{1}$ and $x_{2}$, and to both exclude $\neg x_{1}$ and $\neg x_{2}$  after training. Therefore, once the TM can converge correctly to the intended operation, the actions of $\mathrm{TA}_1$, $\mathrm{TA}_2$, $\mathrm{TA}_3$, and $\mathrm{TA}_4$ should be\footnote{We use ``I" and ``E" as abbreviations for include and exclude respectively.} I, E, I, and E.

To analyze the convergence of TM in the training process, we freeze the transition of the two TAs for the first bit of the input and study behavior of the second bit of input.  Clearly, there are four possibilities for the first bit, $x_{1}$, as:

\begin{itemize}
    \item{Case 1: $\mathrm{TA}_1=\text{E}$, $\mathrm{TA}_2=\text{I}$, i.e., include $\neg x_{1}$}.
    \item{Case 2: $\mathrm{TA}_1=\text{I}$, $\mathrm{TA}_2=\text{E}$, i.e., include $x_{1}$}.
    \item{Case 3: $\mathrm{TA}_1=\text{E}$, $\mathrm{TA}_2=\text{E}$, i.e., exclude both $x_{1}$ and $\neg x_{1}$}.
    \item{Case 4: $\mathrm{TA}_1=\text{I}$, $\mathrm{TA}_2=\text{I}$, i.e., include both $x_{1}$ and $\neg x_{1}$}.
\end{itemize}

In the next subsections, we will analyze the behavior of the TAs for the second bit with different input training samples, given the above four distinct cases of $x_{1}$, respectively.

\subsection{Case 1: Include $\neg x_{1}$}\label{case1}


In this subsection, we assume that the TAs for first bit is frozen as $\mathrm{TA}_{1}=\text{E}$ and $\mathrm{TA}_{2}=\text{I}$, and thus the overall joint action for the first bit is ``$\neg x_{1}$". In this case, we have 4 situations to study, detailed below:

\begin{enumerate}
\item We study the transition of $\mathrm{TA}_{3}$ when it has  ``Include" as its current action, given  different actions of $\mathrm{TA}_{4}$ (i.e., when the action of $\mathrm{TA}_{4}$ is frozen as ``Include" or ``Exclude".). \label {subcase1}
\item We study the transition of $\mathrm{TA}_{3}$ when it has ``Exclude" as its current action, given different actions of $\mathrm{TA}_{4}$ (i.e., when the action of $\mathrm{TA}_{4}$ is frozen as ``Include" or ``Exclude".). \label {subcase2}
\item We study the transition of $\mathrm{TA}_{4}$ when it has ``Include" as its current action, given different actions of $\mathrm{TA}_{3}$ (i.e., when the action of $\mathrm{TA}_{3}$ is frozen as ``Include" or ``Exclude".). \label {subcase3}
\item We study the transition of $\mathrm{TA}_{4}$ when it has ``Exclude" as its current action, given different actions of $\mathrm{TA}_{3}$ (i.e., when the action of $\mathrm{TA}_{3}$ is frozen as ``Include" or ``Exclude".). \label {subcase4}
\end{enumerate}

In what follows, we will go through, exhaustively, the four situations.

\subsubsection{Study $\mathrm{TA}_{3}$ with Action \textsl{Include}}\label{negx1}
Here we study the transitions of TA$_3$ when its current action is \textsl{Include}, given different actions of TA$_4$ and input samples. For ease of expressions, the self-loops of the transitions are not depicted in the transition diagram. 
Clearly, this situation has 8 instances, depending on the variations of the training samples and the status of $\mathrm{TA}_4$, where the first four correspond to the instances with $\mathrm{TA}_4=\text{E}$ while the last four represent the instances with $\mathrm{TA}_4=\text{I}$. 


Now we study the first instance, with $x_{1}=1$, $x_{2}=1$, $y=1$, and $\mathrm{TA}_{4}=\text{E}$. Clearly, this training sample will trigger Type I feedback because $y=1$. Together with the current status of the other TAs, the clause is determined to be $C=  \neg x_{1} \wedge x_{2} = 0$ and the literal is $x_2=1$. From Table \ref{table:type_i_feedback}, we know that the penalty probability is $\frac{1}{s}$ and the inaction probability is $\frac{s-1}{s}$. To indicate the transitions, we have plotted the diagram showing the penalty probability. Note that the overall probability is $u_1\frac{1}{s}$, where $u_1$ is defined in Eq.~(\ref{u1}). To illustrate the transitions of the TA, we assume here we find a certain $T$ such that $u_1>0$ holds.

\begin{minipage}{0.24\textwidth}
Condition:
$x_{1}=1$, $x_{2}=1$, $y=1$, $\mathrm{TA}_{4}=\text{E}$.\\
Thus, Type I, $x_2=1$,    \\
$C= \neg x_{1} \wedge x_{2}=0$.  
\end{minipage}
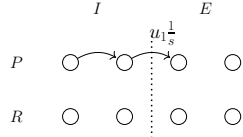
\resizebox{0.18\textwidth}{!}{
\begin{minipage}{0.24\textwidth}
\begin{tikzpicture}[node distance = .35cm, font=\Huge]
    \tikzstyle{every node}=[scale=0.35]
    
    \node[state] (E) at (1,1) {};
    \node[state] (F) at (2,1) {};
    \node[state] (G) at (3,1) {};
    \node[state] (H) at (4,1) {};
  
    \node[state] (A) at (1,2) {};
    \node[state] (B) at (2,2) {};
    \node[state] (C) at (3,2) {};
    \node[state] (D) at (4,2) {};
    
    \node[thick] at (0,1) {$R$};
    \node[thick] at (0,2) {$P$};
    \node[thick] at (1.5,3) {$I$};
    \node[thick] at (3.5,3) {$E$};
    
    \draw[dotted, thick] (2.5,0.5) -- (2.5,2.5);
    
    \draw[every loop]
    (A) edge[bend left] node [scale=1.2, above=0.1 of C] {} (B)
    (B) edge[bend left] node [scale=1.2, above=0.1 of C] {$~~~~u_1\frac{1}{s}$} (C);

\end{tikzpicture}
\end{minipage}
}

We here continue with analyzing another example shown below. In this instance, it covers the training samples: $x_{1}=1$, $x_{2}=0$, $y=0$, and $\mathrm{TA}_{4}=\text{E}$. Clearly, the training sample will trigger Type II feedback because $y=0$. Then the clause becomes $C_{3}= \neg x_{1} \wedge x_{2} = 0$. Because we now study $\mathrm{TA}_3$, the corresponding literal is $x_2=0$. Based on the information above, we can check from Table~\ref{table:type_ii_feedback} that the probability of ``Inaction'' is 1. For this reason, the transition diagram does not have any arrow, indicating that there is ``No transition'' for $\mathrm{TA}_3$.


\begin{minipage}{0.24\textwidth}
Condition:
$x_{1}=1$, $x_{2}=0$, $y=0$, $\mathrm{TA}_4=\text{E}$.\\
Thus, Type II, $x_{2}=0$, \\$C=\neg x_{1}\wedge x_{2}=$0.
\end{minipage}
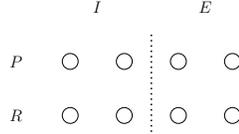
\resizebox{0.18\textwidth}{!}{
\begin{minipage}{0.24\textwidth}
\begin{tikzpicture}[node distance = .35cm, font=\Huge]
    \tikzstyle{every node}=[scale=0.35]
    
    \node[state] (E) at (1,1) {};
    \node[state] (F) at (2,1) {};
    \node[state] (G) at (3,1) {};
    \node[state] (H) at (4,1) {};
  
    \node[state] (A) at (1,2) {};
    \node[state] (B) at (2,2) {};
    \node[state] (C) at (3,2) {};
    \node[state] (D) at (4,2) {};
    
    \node[thick] at (0,1) {$R$};
    \node[thick] at (0,2) {$P$};
    \node[thick] at (1.5,3) {$I$};
    \node[thick] at (3.5,3) {$E$};
    
    \draw[dotted, thick] (2.5,0.5) -- (2.5,2.5);
    

\end{tikzpicture}
\end{minipage}
}
\begin{minipage}{.5\textwidth}
\hspace{6cm}\small{No transition}
\end{minipage}

The same analytical principle applies for all the other instances, and we therefore will not explain them in detail in the remainder of the paper.

\begin{minipage}{0.24\textwidth}
Condition:
$x_{1}=0$, $x_{2}=1$, $y=0$,  $\mathrm{TA}_{4}= $E.\\
Thus, Type II,  $x_{2}=1$,\\ $C=\neg x_{1}\wedge x_{2}=1$.
\end{minipage}
\resizebox{0.18\textwidth}{!}{
\begin{minipage}{0.24\textwidth}
\begin{tikzpicture}[node distance = .35cm, font=\Huge]
    \tikzstyle{every node}=[scale=0.35]
    
    \node[state] (E) at (1,1) {};
    \node[state] (F) at (2,1) {};
    \node[state] (G) at (3,1) {};
    \node[state] (H) at (4,1) {};
  
    \node[state] (A) at (1,2) {};
    \node[state] (B) at (2,2) {};
    \node[state] (C) at (3,2) {};
    \node[state] (D) at (4,2) {};
    
    \node[thick] at (0,1) {$R$};
    \node[thick] at (0,2) {$P$};
    \node[thick] at (1.5,3) {$I$};
    \node[thick] at (3.5,3) {$E$};
    
    \draw[dotted, thick] (2.5,0.5) -- (2.5,2.5);
    

\end{tikzpicture}
\end{minipage}
}
\begin{minipage}{.5\textwidth}
\hspace{6cm}\small{No transition}
\end{minipage}

\begin{minipage}{0.24\textwidth}
Condition:
$x_{1}=0$, $x_{2}=0$, $y=0$, $\mathrm{TA}_{4}= $E.\\
Thus, Type II,  $x_{2}=0$,\\ $C=\neg x_{1}\wedge x_{2}=0$.
\end{minipage}
\resizebox{0.18\textwidth}{!}{
\begin{minipage}{0.24\textwidth}
\begin{tikzpicture}[node distance = .35cm, font=\Huge]
    \tikzstyle{every node}=[scale=0.35]
    
    \node[state] (E) at (1,1) {};
    \node[state] (F) at (2,1) {};
    \node[state] (G) at (3,1) {};
    \node[state] (H) at (4,1) {};
  
    \node[state] (A) at (1,2) {};
    \node[state] (B) at (2,2) {};
    \node[state] (C) at (3,2) {};
    \node[state] (D) at (4,2) {};
    
    \node[thick] at (0,1) {$R$};
    \node[thick] at (0,2) {$P$};
    \node[thick] at (1.5,3) {$I$};
    \node[thick] at (3.5,3) {$E$};
    
    \draw[dotted, thick] (2.5,0.5) -- (2.5,2.5);
    

\end{tikzpicture}
\end{minipage}
}
\begin{minipage}{.5\textwidth}
\hspace{6cm}\small{No transition}
\end{minipage}

\begin{minipage}{0.24\textwidth}
Condition:
$x_{1}=1$, $x_{2}=1$, $y=1$, $\mathrm{TA}_{4}=\text{I}$.\\
Thus, Type I, $x_2=1$, \\
$C= \neg x_{1} \wedge x_{2} \wedge \neg x_{2}=0$.  
\end{minipage}
\hspace{0.6cm}
\resizebox{0.18\textwidth}{!}{
\begin{minipage}{0.24\textwidth}
\begin{tikzpicture}[node distance = .35cm, font=\Huge]
    \tikzstyle{every node}=[scale=0.35]
    
    \node[state] (E) at (1,1) {};
    \node[state] (F) at (2,1) {};
    \node[state] (G) at (3,1) {};
    \node[state] (H) at (4,1) {};
  
    \node[state] (A) at (1,2) {};
    \node[state] (B) at (2,2) {};
    \node[state] (C) at (3,2) {};
    \node[state] (D) at (4,2) {};
    
    \node[thick] at (0,1) {$R$};
    \node[thick] at (0,2) {$P$};
    \node[thick] at (1.5,3) {$I$};
    \node[thick] at (3.5,3) {$E$};
    
    \draw[dotted, thick] (2.5,0.5) -- (2.5,2.5);
    
    \draw[every loop]
    (A) edge[bend left] node [scale=1.2, above=0.1 of C] {} (B)
    (B) edge[bend left] node [scale=1.2, above=0.1 of C] {$~~~~u_1\frac{1}{s}$} (C);

\end{tikzpicture}
\end{minipage}
}

\begin{minipage}{0.24\textwidth}
Condition:
$x_{1}=1$, $x_{2}=0$, $y=0$, $\mathrm{TA}_{4}=\text{I}$. \\
Thus, Type II, $x_{2}=0$, \\
$C=\neg x_{1}\wedge x_{2} \wedge \neg x_{2}=0$.
\end{minipage}
\resizebox{0.18\textwidth}{!}{
\begin{minipage}{0.24\textwidth}
\begin{tikzpicture}[node distance = .35cm, font=\Huge]
    \tikzstyle{every node}=[scale=0.35]
    
    \node[state] (E) at (1,1) {};
    \node[state] (F) at (2,1) {};
    \node[state] (G) at (3,1) {};
    \node[state] (H) at (4,1) {};
  
    \node[state] (A) at (1,2) {};
    \node[state] (B) at (2,2) {};
    \node[state] (C) at (3,2) {};
    \node[state] (D) at (4,2) {};
    
    \node[thick] at (0,1) {$R$};
    \node[thick] at (0,2) {$P$};
    \node[thick] at (1.5,3) {$I$};
    \node[thick] at (3.5,3) {$E$};
    
    \draw[dotted, thick] (2.5,0.5) -- (2.5,2.5);
    

\end{tikzpicture}
\end{minipage}
}
\begin{minipage}{.5\textwidth}
\hspace{6cm}\small{No transition}
\end{minipage}

\begin{minipage}{0.24\textwidth}
Condition:
$x_{1}=0$, $x_{2}=1$, $y=0$, $\mathrm{TA}_{4}=\text{I}$.\\
Thus, Type II, $x_{2}=1$, \\
$C=\neg x_{1}\wedge x_{2} \wedge \neg x_{2}=0$.
\end{minipage}
\resizebox{0.18\textwidth}{!}{
\begin{minipage}{0.24\textwidth}
\begin{tikzpicture}[node distance = .35cm, font=\Huge]
    \tikzstyle{every node}=[scale=0.35]
    
    \node[state] (E) at (1,1) {};
    \node[state] (F) at (2,1) {};
    \node[state] (G) at (3,1) {};
    \node[state] (H) at (4,1) {};
  
    \node[state] (A) at (1,2) {};
    \node[state] (B) at (2,2) {};
    \node[state] (C) at (3,2) {};
    \node[state] (D) at (4,2) {};
    
    \node[thick] at (0,1) {$R$};
    \node[thick] at (0,2) {$P$};
    \node[thick] at (1.5,3) {$I$};
    \node[thick] at (3.5,3) {$E$};
    
    \draw[dotted, thick] (2.5,0.5) -- (2.5,2.5);
    

\end{tikzpicture}
\end{minipage}
}
\begin{minipage}{.5\textwidth}
\hspace{6cm}\small{No transition}
\end{minipage}

\begin{minipage}{0.24\textwidth}
Condition:
$x_{1}=0$, $x_{2}=0$, $y=0$, $\mathrm{TA}_{4}=\text{I}$.\\
Thus, Type II, $x_{2}=0$, \\
$C=\neg x_{1}\wedge x_{2} \wedge \neg x_{2}=0$. 
\end{minipage}
\resizebox{0.18\textwidth}{!}{
\begin{minipage}{0.24\textwidth}
\begin{tikzpicture}[node distance = .35cm, font=\Huge]
    \tikzstyle{every node}=[scale=0.35]
    
    \node[state] (E) at (1,1) {};
    \node[state] (F) at (2,1) {};
    \node[state] (G) at (3,1) {};
    \node[state] (H) at (4,1) {};
  
    \node[state] (A) at (1,2) {};
    \node[state] (B) at (2,2) {};
    \node[state] (C) at (3,2) {};
    \node[state] (D) at (4,2) {};
    
    \node[thick] at (0,1) {$R$};
    \node[thick] at (0,2) {$P$};
    \node[thick] at (1.5,3) {$I$};
    \node[thick] at (3.5,3) {$E$};
    
    \draw[dotted, thick] (2.5,0.5) -- (2.5,2.5);
    

\end{tikzpicture}
\end{minipage}
}
\begin{minipage}{.5\textwidth}
\hspace{6cm}\small{No transition}
\end{minipage}

\subsubsection{Study $\mathrm{TA}_{3}$ with Action \textsl{Exclude}}
Here we study the transitions of TA$_3$ when its current action is \textsl{Exclude}, given different actions of TA$_4$ and input samples. This situation has 8 instances, depending on the variations of the training samples and the status of $\mathrm{TA}_4$. In this subsection and the following subsections, we will not plot the transition diagrams for ``No transition". 

\begin{minipage}{0.24\textwidth}
Condition:
$x_{1}=1$, $x_{2}=1$, $y=1$, $\mathrm{TA}_{4}=\text{E}$.\\
Thus, Type I, $x_{2}=1$,\\
$C= \neg x_{1}=0$.  
\end{minipage}
\resizebox{0.18\textwidth}{!}{
\begin{minipage}{0.24\textwidth}
\begin{tikzpicture}[node distance = .35cm, font=\Huge]
    \tikzstyle{every node}=[scale=0.35]
    
    \node[state] (E) at (1,1) {};
    \node[state] (F) at (2,1) {};
    \node[state] (G) at (3,1) {};
    \node[state] (H) at (4,1) {};
  
    \node[state] (A) at (1,2) {};
    \node[state] (B) at (2,2) {};
    \node[state] (C) at (3,2) {};
    \node[state] (D) at (4,2) {};
    
    \node[thick] at (0,1) {$R$};
    \node[thick] at (0,2) {$P$};
    \node[thick] at (1.5,3) {$I$};
    \node[thick] at (3.5,3) {$E$};
    
    \draw[dotted, thick] (2.5,0.5) -- (2.5,2.5);
    
    \draw[every loop]
    (G) edge[bend left] node [scale=1.2, above=0.1 of C] {} (H)
    (H) edge[loop right] node [scale=1.2, above=0.1 of H] {$u_1\frac{1}{s}$} (H);

\end{tikzpicture}
\end{minipage}
}

\begin{minipage}{0.24\textwidth}
Condition:
$x_{1}=0$, $x_{2}=0$, $y=0$, $\mathrm{TA}_{4}=\text{E}$.\\
Thus, Type II, $x_2=0$,  \\
$C=\neg x_{1}=1$.  
\end{minipage}
\resizebox{0.18\textwidth}{!}{
\begin{minipage}{0.24\textwidth}
\begin{tikzpicture}[node distance = .35cm, font=\Huge]
    \tikzstyle{every node}=[scale=0.35]
    
    \node[state] (E) at (1,1) {};
    \node[state] (F) at (2,1) {};
    \node[state] (G) at (3,1) {};
    \node[state] (H) at (4,1) {};
  
    \node[state] (A) at (1,2) {};
    \node[state] (B) at (2,2) {};
    \node[state] (C) at (3,2) {};
    \node[state] (D) at (4,2) {};
    
    \node[thick] at (0,1) {$R$};
    \node[thick] at (0,2) {$P$};
    \node[thick] at (1.5,3) {$I$};
    \node[thick] at (3.5,3) {$E$};
    
    \draw[dotted, thick] (2.5,0.5) -- (2.5,2.5);
    
    \draw[every loop]
    (D) edge[bend right] node [scale=1.2, above=0.1 of C] {$u_2\times1$} (C)
    (C) edge[bend right] node [scale=1.2, above=0.1 of B] {} (B);

\end{tikzpicture}
\end{minipage}
}

\begin{minipage}{0.24\textwidth}
Condition:
$x_{1}=1$, $x_{2}=1$, $y=1$,  $\mathrm{TA}_{4}=\text{I}$. \\
Thus, Type I, $x_2=1$,  \\
$C=\neg x_{1} \wedge \neg x_{2}=0$.  
\end{minipage}
\resizebox{0.18\textwidth}{!}{
\begin{minipage}{0.24\textwidth}
\begin{tikzpicture}[node distance = .35cm, font=\Huge]
    \tikzstyle{every node}=[scale=0.35]
    
    \node[state] (E) at (1,1) {};
    \node[state] (F) at (2,1) {};
    \node[state] (G) at (3,1) {};
    \node[state] (H) at (4,1) {};
  
    \node[state] (A) at (1,2) {};
    \node[state] (B) at (2,2) {};
    \node[state] (C) at (3,2) {};
    \node[state] (D) at (4,2) {};
    
    \node[thick] at (0,1) {$R$};
    \node[thick] at (0,2) {$P$};
    \node[thick] at (1.5,3) {$I$};
    \node[thick] at (3.5,3) {$E$};
    
    \draw[dotted, thick] (2.5,0.5) -- (2.5,2.5);
    
    \draw[every loop]
    (G) edge[bend left] node [scale=1.2, above=0.1 of C] {} (H)
    (H) edge[loop right] node [scale=1.2, below=0.1 of H] {$u_1\frac{1}{s}$} (H);

\end{tikzpicture}
\end{minipage}
}

\begin{minipage}{0.24\textwidth}
Condition:
$x_{1}=0$, $x_{2}=0$, $y=0$, $\mathrm{TA}_{4}=\text{I}$.\\
Thus, Type II, $x_2=0$, \\
$C=\neg x_{1} \wedge \neg x_{2}=1$.  
\end{minipage}
\resizebox{0.18\textwidth}{!}{
\begin{minipage}{0.24\textwidth}
\begin{tikzpicture}[node distance = .35cm, font=\Huge]
    \tikzstyle{every node}=[scale=0.35]
    
    \node[state] (E) at (1,1) {};
    \node[state] (F) at (2,1) {};
    \node[state] (G) at (3,1) {};
    \node[state] (H) at (4,1) {};
  
    \node[state] (A) at (1,2) {};
    \node[state] (B) at (2,2) {};
    \node[state] (C) at (3,2) {};
    \node[state] (D) at (4,2) {};
    
    \node[thick] at (0,1) {$R$};
    \node[thick] at (0,2) {$P$};
    \node[thick] at (1.5,3) {$I$};
    \node[thick] at (3.5,3) {$E$};
    
    \draw[dotted, thick] (2.5,0.5) -- (2.5,2.5);
    
    \draw[every loop]
    (D) edge[bend right] node [scale=1.2, above=0.1 of C] {$u_2\times1$} (C)
    (C) edge[bend right] node [scale=1.2, above=0.1 of B] {} (B);

\end{tikzpicture}
\end{minipage}
}

\subsubsection{Study $\mathrm{TA}_{4}$ with Action \textsl{Include}}
\hspace{0.55cm}\begin{minipage}{0.24\textwidth}
Condition:
$x_{1}=1$, $x_{2}=1$, $y=1$,  $\mathrm{TA}_{3}=\text{E}$.\\
Thus, Type I, $\neg x_{2} = 0$, \\
$C=\neg x_{1} \wedge \neg x_{2}=0$.  
\end{minipage}
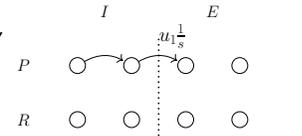
\resizebox{0.18\textwidth}{!}{
\begin{minipage}{0.24\textwidth}
\begin{tikzpicture}[node distance = .35cm, font=\Huge]
    \tikzstyle{every node}=[scale=0.35]
    
    \node[state] (E) at (1,1) {};
    \node[state] (F) at (2,1) {};
    \node[state] (G) at (3,1) {};
    \node[state] (H) at (4,1) {};
  
    \node[state] (A) at (1,2) {};
    \node[state] (B) at (2,2) {};
    \node[state] (C) at (3,2) {};
    \node[state] (D) at (4,2) {};
    
    \node[thick] at (0,1) {$R$};
    \node[thick] at (0,2) {$P$};
    \node[thick] at (1.5,3) {$I$};
    \node[thick] at (3.5,3) {$E$};
    
    \draw[dotted, thick] (2.5,0.5) -- (2.5,2.5);
    
    \draw[every loop]
    (A) edge[bend left] node [scale=1.2, above=0.1 of C] {} (B)
    (B) edge[bend left] node [scale=1.2, above=0.1 of C] {$~~~~~u_1\frac{1}{s}$} (C);

\end{tikzpicture}
\end{minipage}
}

\begin{minipage}{0.24\textwidth}
Condition:
$x_{1}=1$, $x_{2}=1$, $y=1$, $\mathrm{TA}_{3}=\text{I}$.\\
Thus, Type I, $\neg x_{2}=0$, \\
$C=\neg x_{1} \wedge x_{2} \wedge \neg x_{2}=0$.  
\end{minipage}
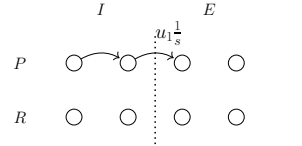
\resizebox{0.18\textwidth}{!}{
\begin{minipage}{0.24\textwidth}
\begin{tikzpicture}[node distance = .35cm, font=\Huge]
    \tikzstyle{every node}=[scale=0.35]
    
    \node[state] (E) at (1,1) {};
    \node[state] (F) at (2,1) {};
    \node[state] (G) at (3,1) {};
    \node[state] (H) at (4,1) {};
  
    \node[state] (A) at (1,2) {};
    \node[state] (B) at (2,2) {};
    \node[state] (C) at (3,2) {};
    \node[state] (D) at (4,2) {};
    
    \node[thick] at (0,1) {$R$};
    \node[thick] at (0,2) {$P$};
    \node[thick] at (1.5,3) {$I$};
    \node[thick] at (3.5,3) {$E$};
    
    \draw[dotted, thick] (2.5,0.5) -- (2.5,2.5);
    
    \draw[every loop]
    (A) edge[bend left] node [scale=1.2, above=0.1 of C] {} (B)
    (B) edge[bend left] node [scale=1.2, above=0.1 of H] {$~~~~~u_1\frac{1}{s}$} (C);

\end{tikzpicture}
\end{minipage}
}

\subsubsection{Study $\mathrm{TA}_{4}$ with Action \textsl{Exclude}}\label{ExcludeTA4}
\hspace{0.55cm}\begin{minipage}{0.24\textwidth}
Condition:
$x_{1}=1$, $x_{2}=1$, $y=1$, $\mathrm{TA}_{3}=\text{E}$.\\
Thus, Type I, $\neg x_{2}=0$, \\
$C=\neg x_{1}=0$.  
\end{minipage}
\resizebox{0.18\textwidth}{!}{
\begin{minipage}{0.24\textwidth}
\begin{tikzpicture}[node distance = .35cm, font=\Huge]
    \tikzstyle{every node}=[scale=0.35]
    
    \node[state] (E) at (1,1) {};
    \node[state] (F) at (2,1) {};
    \node[state] (G) at (3,1) {};
    \node[state] (H) at (4,1) {};
  
    \node[state] (A) at (1,2) {};
    \node[state] (B) at (2,2) {};
    \node[state] (C) at (3,2) {};
    \node[state] (D) at (4,2) {};
    
    \node[thick] at (0,1) {$R$};
    \node[thick] at (0,2) {$P$};
    \node[thick] at (1.5,3) {$I$};
    \node[thick] at (3.5,3) {$E$};
    
    \draw[dotted, thick] (2.5,0.5) -- (2.5,2.5);
    
    \draw[every loop]
    (G) edge[bend left] node [scale=1.2, above=0.1 of C] {} (H)
    (H) edge[loop right] node [scale=1.2, below=0.1 of H] {$u_1\frac{1}{s}$} (H);

\end{tikzpicture}
\end{minipage}
}

\begin{minipage}{0.24\textwidth}
Condition:
$x_{1}=0$, $x_{2}=1$, $y=0$, $\mathrm{TA}_{3}=\text{E}$.\\
Thus, Type II, $\neg x_{2}=0$, \\
$C=\neg x_{1}=1$.  
\end{minipage}
\resizebox{0.18\textwidth}{!}{
\begin{minipage}{0.24\textwidth}
\begin{tikzpicture}[node distance = .35cm, font=\Huge]
    \tikzstyle{every node}=[scale=0.35]
    
    \node[state] (E) at (1,1) {};
    \node[state] (F) at (2,1) {};
    \node[state] (G) at (3,1) {};
    \node[state] (H) at (4,1) {};
  
    \node[state] (A) at (1,2) {};
    \node[state] (B) at (2,2) {};
    \node[state] (C) at (3,2) {};
    \node[state] (D) at (4,2) {};
    
    \node[thick] at (0,1) {$R$};
    \node[thick] at (0,2) {$P$};
    \node[thick] at (1.5,3) {$I$};
    \node[thick] at (3.5,3) {$E$};
    
    \draw[dotted, thick] (2.5,0.5) -- (2.5,2.5);
    
    \draw[every loop]
    (D) edge[bend right] node [scale=1.2, above=0.1 of C] {$u_2\times1$} (C)
    (C) edge[bend right] node [scale=1.2, above=0.1 of B] {} (B);

\end{tikzpicture}
\end{minipage}
}

\begin{minipage}{0.24\textwidth}
Condition:
$x_{1}=1$, $x_{2}=1$, $y=1$, $\mathrm{TA}_{3}=\text{I}$.\\
Thus, Type I, $\neg x_{2}=0$, \\
$C=\neg x_{1} \wedge x_{2}=0$.  
\end{minipage}
\resizebox{0.18\textwidth}{!}{
\begin{minipage}{0.24\textwidth}
\begin{tikzpicture}[node distance = .35cm, font=\Huge]
    \tikzstyle{every node}=[scale=0.35]
    
    \node[state] (E) at (1,1) {};
    \node[state] (F) at (2,1) {};
    \node[state] (G) at (3,1) {};
    \node[state] (H) at (4,1) {};
  
    \node[state] (A) at (1,2) {};
    \node[state] (B) at (2,2) {};
    \node[state] (C) at (3,2) {};
    \node[state] (D) at (4,2) {};
    
    \node[thick] at (0,1) {$R$};
    \node[thick] at (0,2) {$P$};
    \node[thick] at (1.5,3) {$I$};
    \node[thick] at (3.5,3) {$E$};
    
    \draw[dotted, thick] (2.5,0.5) -- (2.5,2.5);
    
    \draw[every loop]
    (G) edge[bend left] node [scale=1.2, above=0.1 of C] {} (H)
    (H) edge[loop right] node [scale=1.2, below=0.1 of H] {$u_1\frac{1}{s}$} (H);

\end{tikzpicture}
\end{minipage}
}

\begin{minipage}{0.24\textwidth}
Condition:
$x_{1}=0$, $x_{2}=1$, $y=0$, $\mathrm{TA}_{3}=\text{I}$.\\
Thus, Type II, $\neg x_{2}=0$, \\
$C=\neg x_{1} \wedge x_{2}=1$.  
\end{minipage}
\resizebox{0.18\textwidth}{!}{
\begin{minipage}{0.24\textwidth}
\begin{tikzpicture}[node distance = .35cm, font=\Huge]
    \tikzstyle{every node}=[scale=0.35]
    
    \node[state] (E) at (1,1) {};
    \node[state] (F) at (2,1) {};
    \node[state] (G) at (3,1) {};
    \node[state] (H) at (4,1) {};
  
    \node[state] (A) at (1,2) {};
    \node[state] (B) at (2,2) {};
    \node[state] (C) at (3,2) {};
    \node[state] (D) at (4,2) {};
    
    \node[thick] at (0,1) {$R$};
    \node[thick] at (0,2) {$P$};
    \node[thick] at (1.5,3) {$I$};
    \node[thick] at (3.5,3) {$E$};
    
    \draw[dotted, thick] (2.5,0.5) -- (2.5,2.5);
    
    \draw[every loop]
    (D) edge[bend right] node [scale=1.2, above=0.1 of C] {$u_2\times1$} (C)
    (C) edge[bend right] node [scale=1.2, above=0.1 of B] {} (B);

\end{tikzpicture}
\end{minipage}
}

\subsection{Case 2: Include $x_{1}$}\label{case2}

For Case 2, we assume that the actions of the TAs for the first bit are frozen as $\mathrm{TA}_{1}=\text{I}$ and $\mathrm{TA}_{2}=\text{E}$, and thus the overall joint action for the first bit is ``$x_{1}$". Similar to Case 1, we also have 4 situations, detailed in subsections \ref{x1}-\ref{x1exclude}. 

\subsubsection{Study $\mathrm{TA}_{3}$ with Action \textsl{Include}}
\label{x1}

\hspace{0.55cm}\begin{minipage}{0.24\textwidth}
Condition:
$x_{1}=1$, $x_{2}=1$, $y=1$, $\mathrm{TA}_{4}=\text{E}$.\\
Thus, Type I, $x_2=1$,\\ $C=x_{1}\wedge x_{2}$=1.
\end{minipage}
\resizebox{0.18\textwidth}{!}{
\begin{minipage}{0.24\textwidth}
\begin{tikzpicture}[node distance = .35cm, font=\Huge]
    \tikzstyle{every node}=[scale=0.35]
    
    \node[state] (E) at (1,1) {};
    \node[state] (F) at (2,1) {};
    \node[state] (G) at (3,1) {};
    \node[state] (H) at (4,1) {};
  
    \node[state] (A) at (1,2) {};
    \node[state] (B) at (2,2) {};
    \node[state] (C) at (3,2) {};
    \node[state] (D) at (4,2) {};
    
    \node[thick] at (0,1) {$R$};
    \node[thick] at (0,2) {$P$};
    \node[thick] at (1.5,3) {$I$};
    \node[thick] at (3.5,3) {$E$};
    
    \draw[dotted, thick] (2.5,0.5) -- (2.5,2.5);
    
    \draw[every loop]
    (F) edge[bend left] node {} (E)
    (E) edge[loop left=45] node [scale=1.2, below=0.1 of E] {$u_1\frac{s-1}{s}$} (E);

\end{tikzpicture}
\end{minipage}
}

\begin{minipage}{0.24\textwidth}
Condition:
$x_{1}=1$, $x_{2}=1$, $y=1$,  $\mathrm{TA}_{4}=\text{I}$.\\
Thus, Type I,  $x_{2}=1$,\\
$C=x_{1}\wedge x_{2} \wedge \neg x_{2}=0$.
\end{minipage}
\resizebox{0.18\textwidth}{!}{
\begin{minipage}{0.24\textwidth}
\begin{tikzpicture}[node distance = .35cm, font=\Huge]
    \tikzstyle{every node}=[scale=0.35]
    
    \node[state] (E) at (1,1) {};
    \node[state] (F) at (2,1) {};
    \node[state] (G) at (3,1) {};
    \node[state] (H) at (4,1) {};
  
    \node[state] (A) at (1,2) {};
    \node[state] (B) at (2,2) {};
    \node[state] (C) at (3,2) {};
    \node[state] (D) at (4,2) {};
    
    \node[thick] at (0,1) {$R$};
    \node[thick] at (0,2) {$P$};
    \node[thick] at (1.5,3) {$I$};
    \node[thick] at (3.5,3) {$E$};
    
    \draw[dotted, thick] (2.5,0.5) -- (2.5,2.5);
    
    \draw[every loop]
    (A) edge[bend left] node {} (B)
    (B) edge[bend left] node [scale=1.2, above=0.1 of E] {$~~~~~~u_1\frac{1}{s}$} (C);

\end{tikzpicture}
\end{minipage}
}

\subsubsection{Study $\mathrm{TA}_{3}$ with Action \textsl{Exclude}}

\hspace{0.55cm}\begin{minipage}{0.24\textwidth}
Condition:
$x_{1}=1$, $x_{2}=1$, $y=1$,  $\mathrm{TA}_{4}=\text{E}$.\\
Thus, Type I, $x_{2}=1$, \\
$C=x_{1}$=1. 
\end{minipage}
\resizebox{0.18\textwidth}{!}{
\begin{minipage}{0.24\textwidth}
\begin{tikzpicture}[node distance = .35cm, font=\Huge]
    \tikzstyle{every node}=[scale=0.35]
    
    \node[state] (E) at (1,1) {};
    \node[state] (F) at (2,1) {};
    \node[state] (G) at (3,1) {};
    \node[state] (H) at (4,1) {};
  
    \node[state] (A) at (1,2) {};
    \node[state] (B) at (2,2) {};
    \node[state] (C) at (3,2) {};
    \node[state] (D) at (4,2) {};
    
    \node[thick] at (0,1) {$R$};
    \node[thick] at (0,2) {$P$};
    \node[thick] at (1.5,3) {$I$};
    \node[thick] at (3.5,3) {$E$};
    
    \draw[dotted, thick] (2.5,0.5) -- (2.5,2.5);
    
    \draw[every loop]
    (D) edge[bend right] node [scale=1.2, above=0.1 of C] {$~~u_1\frac{s-1}{s}$} (C)
    (C) edge[bend right] node [scale=1.2, above=0.1 of B] {} (B);

\end{tikzpicture}
\end{minipage}
}

\begin{minipage}{0.24\textwidth}
Condition:
$x_{1}=1$, $x_{2}=0$, $y=0$,  $\mathrm{TA}_{4}=\text{E}$.\\
Thus, Type II, $x_{2}=0$, \\
$C=x_{1}=1$.
\end{minipage}
\resizebox{0.18\textwidth}{!}{
\begin{minipage}{0.24\textwidth}
\begin{tikzpicture}[node distance = .35cm, font=\Huge]
    \tikzstyle{every node}=[scale=0.35]
    
    \node[state] (E) at (1,1) {};
    \node[state] (F) at (2,1) {};
    \node[state] (G) at (3,1) {};
    \node[state] (H) at (4,1) {};
  
    \node[state] (A) at (1,2) {};
    \node[state] (B) at (2,2) {};
    \node[state] (C) at (3,2) {};
    \node[state] (D) at (4,2) {};
    
    \node[thick] at (0,1) {$R$};
    \node[thick] at (0,2) {$P$};
    \node[thick] at (1.5,3) {$I$};
    \node[thick] at (3.5,3) {$E$};
    
    \draw[dotted, thick] (2.5,0.5) -- (2.5,2.5);
    
    \draw[every loop]
    (D) edge[bend right] node [scale=1.2, above=0.1 of C] {$u_2\times1$} (C)
    (C) edge[bend right] node [scale=1.2, above=0.1 of B] {} (B);

\end{tikzpicture}
\end{minipage}
}

\begin{minipage}{0.24\textwidth}
Condition:
$x_{1}=1$, $x_{2}=1$, $y=1$,  $\mathrm{TA}_{4}=\text{I}$. \\
Thus, Type I, $x_{2}=1$, \\
$C=x_{1} \wedge \neg x_{2}=0$.
\end{minipage}
\resizebox{0.18\textwidth}{!}{
\begin{minipage}{0.24\textwidth}
\begin{tikzpicture}[node distance = .35cm, font=\Huge]
    \tikzstyle{every node}=[scale=0.35]
    
    \node[state] (E) at (1,1) {};
    \node[state] (F) at (2,1) {};
    \node[state] (G) at (3,1) {};
    \node[state] (H) at (4,1) {};
  
    \node[state] (A) at (1,2) {};
    \node[state] (B) at (2,2) {};
    \node[state] (C) at (3,2) {};
    \node[state] (D) at (4,2) {};
    
    \node[thick] at (0,1) {$R$};
    \node[thick] at (0,2) {$P$};
    \node[thick] at (1.5,3) {$I$};
    \node[thick] at (3.5,3) {$E$};
    
    \draw[dotted, thick] (2.5,0.5) -- (2.5,2.5);
    
    \draw[every loop]
    (G) edge[bend left] node [scale=1.2, above=0.1 of C] {} (H)
    (H) edge[loop right] node [scale=1.2, below=0.1 of H] {$u_1\frac{1}{s}$} (H);

\end{tikzpicture}
\end{minipage}
}

\begin{minipage}{0.24\textwidth}
Condition:
$x_{1}=1$, $x_{2}=0$, $y=0$,  $\mathrm{TA}_{4}=\text{I}$.\\
Thus, Type II, $x_2=0$, \\
$C=x_{1} \wedge \neg x_{2}=1$.
\end{minipage}
\resizebox{0.18\textwidth}{!}{
\begin{minipage}{0.24\textwidth}
\begin{tikzpicture}[node distance = .35cm, font=\Huge]
    \tikzstyle{every node}=[scale=0.35]
    
    \node[state] (E) at (1,1) {};
    \node[state] (F) at (2,1) {};
    \node[state] (G) at (3,1) {};
    \node[state] (H) at (4,1) {};
  
    \node[state] (A) at (1,2) {};
    \node[state] (B) at (2,2) {};
    \node[state] (C) at (3,2) {};
    \node[state] (D) at (4,2) {};
    
    \node[thick] at (0,1) {$R$};
    \node[thick] at (0,2) {$P$};
    \node[thick] at (1.5,3) {$I$};
    \node[thick] at (3.5,3) {$E$};
    
    \draw[dotted, thick] (2.5,0.5) -- (2.5,2.5);
    
    \draw[every loop]
    (D) edge[bend right] node [scale=1.2, above=0.1 of C] {$u_2\times1$} (C)
    (C) edge[bend right] node [scale=1.2, below=0.1 of H] {} (B);

\end{tikzpicture}
\end{minipage}
}

\subsubsection{Study $\mathrm{TA}_{4}$ with Action \textsl{Include}}

\hspace{0.55cm}\begin{minipage}{0.24\textwidth}
Condition:
$x_{1}=1$, $x_{2}=1$, $y=1$,  $\mathrm{TA}_{3}=\text{E}$.\\
Thus, Type I, $\neg x_{2}=0$, \\$C=x_{1} \wedge \neg x_{2}=0$.
\end{minipage}
\resizebox{0.18\textwidth}{!}{ 
\begin{minipage}{0.24\textwidth}
\begin{tikzpicture}[node distance = .35cm, font=\Huge]
    \tikzstyle{every node}=[scale=0.35]
    
    \node[state] (E) at (1,1) {};
    \node[state] (F) at (2,1) {};
    \node[state] (G) at (3,1) {};
    \node[state] (H) at (4,1) {};
  
    \node[state] (A) at (1,2) {};
    \node[state] (B) at (2,2) {};
    \node[state] (C) at (3,2) {};
    \node[state] (D) at (4,2) {};
    
    \node[thick] at (0,1) {$R$};
    \node[thick] at (0,2) {$P$};
    \node[thick] at (1.5,3) {$I$};
    \node[thick] at (3.5,3) {$E$};
    
    \draw[dotted, thick] (2.5,0.5) -- (2.5,2.5);
    
    \draw[every loop]
    (A) edge[bend left] node [scale=1.2, above=0.1 of C] {} (B)
    (B) edge[bend left] node [scale=1.2, above=0.1 of C] {$~~~~~u_1\frac{1}{s}$} (C);

\end{tikzpicture}
\end{minipage}
}

\begin{minipage}{0.24\textwidth}
Condition:
$x_{1}=1$, $x_{2}=1$, $y=1$, $\mathrm{TA}_{3}=\text{I}$.\\
Thus, Type I, $\neg x_{2}=0$, \\
$C=x_{1} \wedge x_{2} \wedge\neg x_{2}=0$.
\end{minipage}
\resizebox{0.18\textwidth}{!}{
\begin{minipage}{0.24\textwidth}
\begin{tikzpicture}[node distance = .35cm, font=\Huge]
    \tikzstyle{every node}=[scale=0.35]
    
    \node[state] (E) at (1,1) {};
    \node[state] (F) at (2,1) {};
    \node[state] (G) at (3,1) {};
    \node[state] (H) at (4,1) {};
  
    \node[state] (A) at (1,2) {};
    \node[state] (B) at (2,2) {};
    \node[state] (C) at (3,2) {};
    \node[state] (D) at (4,2) {};
    
    \node[thick] at (0,1) {$R$};
    \node[thick] at (0,2) {$P$};
    \node[thick] at (1.5,3) {$I$};
    \node[thick] at (3.5,3) {$E$};
    
    \draw[dotted, thick] (2.5,0.5) -- (2.5,2.5);
    
    \draw[every loop]
    (A) edge[bend left] node [scale=1.2, above=0.1 of C] {} (B)
    (B) edge[bend left] node [scale=1.2, above=0.1 of C] {$~~~~~u_1\frac{1}{s}$} (C);

\end{tikzpicture}
\end{minipage}
}

\subsubsection{Study $\mathrm{TA}_{4}$ with Action \textsl{Exclude}}\label{x1exclude}
\hspace{0.55cm}\begin{minipage}{0.24\textwidth}
Condition:
$x_{1}=1$, $x_{2}=1$, $y=1$, $\mathrm{TA}_{3}=\text{E}$.\\
Thus, Type I, $\neg x_{2}=0$, \\
$C=x_{1}=1$.  
\end{minipage}
\resizebox{0.18\textwidth}{!}{
\begin{minipage}{0.24\textwidth}
\begin{tikzpicture}[node distance = .35cm, font=\Huge]
    \tikzstyle{every node}=[scale=0.35]
    
    \node[state] (E) at (1,1) {};
    \node[state] (F) at (2,1) {};
    \node[state] (G) at (3,1) {};
    \node[state] (H) at (4,1) {};
  
    \node[state] (A) at (1,2) {};
    \node[state] (B) at (2,2) {};
    \node[state] (C) at (3,2) {};
    \node[state] (D) at (4,2) {};
    
    \node[thick] at (0,1) {$R$};
    \node[thick] at (0,2) {$P$};
    \node[thick] at (1.5,3) {$I$};
    \node[thick] at (3.5,3) {$E$};
    
    \draw[dotted, thick] (2.5,0.5) -- (2.5,2.5);
    
    \draw[every loop]
    (G) edge[bend left] node [scale=1.2, above=0.1 of C] {} (H)
    (H) edge[loop right] node [scale=1.2, below=0.1 of H] {$u_1\frac{1}{s}$} (H);

\end{tikzpicture}
\end{minipage}
}

\begin{minipage}{0.24\textwidth}
Condition:
$x_{1}=1$, $x_{2}=1$, $y=1$,  $\mathrm{TA}_{3}=\text{I}$.\\
Thus, Type I, $\neg x_{2}=0$, \\
$C=x_{1} \wedge x_{2}=1$.
\end{minipage}
\resizebox{0.18\textwidth}{!}{
\begin{minipage}{0.24\textwidth}
\begin{tikzpicture}[node distance = .35cm, font=\Huge]
    \tikzstyle{every node}=[scale=0.35]
    
    \node[state] (E) at (1,1) {};
    \node[state] (F) at (2,1) {};
    \node[state] (G) at (3,1) {};
    \node[state] (H) at (4,1) {};
  
    \node[state] (A) at (1,2) {};
    \node[state] (B) at (2,2) {};
    \node[state] (C) at (3,2) {};
    \node[state] (D) at (4,2) {};
    
    \node[thick] at (0,1) {$R$};
    \node[thick] at (0,2) {$P$};
    \node[thick] at (1.5,3) {$I$};
    \node[thick] at (3.5,3) {$E$};
    
    \draw[dotted, thick] (2.5,0.5) -- (2.5,2.5);
    
    \draw[every loop]
    (G) edge[bend left] node [scale=1.2, above=0.1 of C] {} (H)
    (H) edge[loop right] node [scale=1.2, below=0.1 of H] {$u_1\frac{1}{s}$} (H);

\end{tikzpicture}
\end{minipage}
}

\subsection{Case 3: Exclude Both $\neg x_{1}$ and $x_{1}$}\label{case3}

For Case 3, we assume that the actions of TAs for the first bit are frozen as $\mathrm{TA}_{1}=\text{E}$ and $\mathrm{TA}_{2}=\text{E}$, with 4 situations, detailed in subsections \ref{341}-\ref{344}. Note that in the training process, when all literals are excluded, $C$ is assigned to 1. 

\subsubsection{Study $\mathrm{TA}_{3}$ with Action \textsl{Include}}\label{341}

\hspace{0.55cm}\begin{minipage}{0.24\textwidth}
Condition:
$x_{1}=1$, $x_{2}=1$, $y=1$, $\mathrm{TA}_{4}=\text{E}$.\\
Thus, Type I, $x_2=1$, \\
$C=x_{2}=1$.  
\end{minipage}
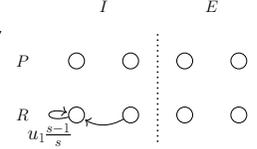
\resizebox{0.18\textwidth}{!}{
\begin{minipage}{0.24\textwidth}
\begin{tikzpicture}[node distance = .35cm, font=\Huge]
    \tikzstyle{every node}=[scale=0.35]
    
    \node[state] (E) at (1,1) {};
    \node[state] (F) at (2,1) {};
    \node[state] (G) at (3,1) {};
    \node[state] (H) at (4,1) {};
  
    \node[state] (A) at (1,2) {};
    \node[state] (B) at (2,2) {};
    \node[state] (C) at (3,2) {};
    \node[state] (D) at (4,2) {};
    
    \node[thick] at (0,1) {$R$};
    \node[thick] at (0,2) {$P$};
    \node[thick] at (1.5,3) {$I$};
    \node[thick] at (3.5,3) {$E$};
    
    \draw[dotted, thick] (2.5,0.5) -- (2.5,2.5);
    
    \draw[every loop]
    (F) edge[bend left] node [scale=1.2, above=0.1 of C] {} (E)
    (E) edge[loop left] node [scale=1.2, below=0.1 of E] {$u_1\frac{s-1}{s}$} (E);

\end{tikzpicture}
\end{minipage}
}

\begin{minipage}{0.24\textwidth}
Condition:
$x_{1}=1$, $x_{2}=1$, $y=1$,  $\mathrm{TA}_{4}=\text{I}$. \\
Thus, Type I, $ x_{2}=1$, \\
$C=0$.  
\end{minipage}
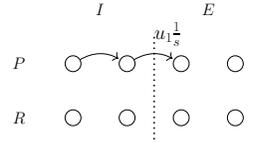
\resizebox{0.18\textwidth}{!}{
\begin{minipage}{0.24\textwidth}
\begin{tikzpicture}[node distance = .35cm, font=\Huge]
    \tikzstyle{every node}=[scale=0.35]
    
    \node[state] (E) at (1,1) {};
    \node[state] (F) at (2,1) {};
    \node[state] (G) at (3,1) {};
    \node[state] (H) at (4,1) {};
  
    \node[state] (A) at (1,2) {};
    \node[state] (B) at (2,2) {};
    \node[state] (C) at (3,2) {};
    \node[state] (D) at (4,2) {};
    
    \node[thick] at (0,1) {$R$};
    \node[thick] at (0,2) {$P$};
    \node[thick] at (1.5,3) {$I$};
    \node[thick] at (3.5,3) {$E$};
    
    \draw[dotted, thick] (2.5,0.5) -- (2.5,2.5);
    
    \draw[every loop]
    (A) edge[bend left] node [scale=1.2, above=0.1 of C] {} (B)
    (B) edge[bend left] node [scale=1.2, above=0.1 of C] {$~~~~~u_1\frac{1}{s}$} (C);

\end{tikzpicture}
\end{minipage}
}

\subsubsection{Study $\mathrm{TA}_{3}$ with Action \textsl{Exclude}}

\hspace{0.55cm}\begin{minipage}{0.24\textwidth}
Condition:
$x_{1}=1$, $x_{2}=1$, $y=1$,  $\mathrm{TA}_{4}=\text{E}$.\\
Thus, Type I, $x_{2}=1$, \\
$C=1$. 
\end{minipage}
\resizebox{0.18\textwidth}{!}{
\begin{minipage}{0.24\textwidth}
\begin{tikzpicture}[node distance = .35cm, font=\Huge]
    \tikzstyle{every node}=[scale=0.35]
    
    \node[state] (E) at (1,1) {};
    \node[state] (F) at (2,1) {};
    \node[state] (G) at (3,1) {};
    \node[state] (H) at (4,1) {};
  
    \node[state] (A) at (1,2) {};
    \node[state] (B) at (2,2) {};
    \node[state] (C) at (3,2) {};
    \node[state] (D) at (4,2) {};
    
    \node[thick] at (0,1) {$R$};
    \node[thick] at (0,2) {$P$};
    \node[thick] at (1.5,3) {$I$};
    \node[thick] at (3.5,3) {$E$};
    
    \draw[dotted, thick] (2.5,0.5) -- (2.5,2.5);
    
    \draw[every loop]
    (D) edge[bend right] node [scale=1.2, above=0.1 of C] {$~~u_1\frac{s-1}{s}$} (C)
    (C) edge[bend right] node [scale=1.2, above=0.1 of B] {} (B);

\end{tikzpicture}
\end{minipage}
}

\begin{minipage}{0.24\textwidth}
Condition:
$x_{1}=1$, $x_{2}=0$, $y=0$, $\mathrm{TA}_{4}=\text{E}$.\\
Thus, Type II, $x_2=0$,  \\
$C=1$.  
\end{minipage}
\resizebox{0.18\textwidth}{!}{
\begin{minipage}{0.24\textwidth}
\begin{tikzpicture}[node distance = .35cm, font=\Huge]
    \tikzstyle{every node}=[scale=0.35]
    
    \node[state] (E) at (1,1) {};
    \node[state] (F) at (2,1) {};
    \node[state] (G) at (3,1) {};
    \node[state] (H) at (4,1) {};
  
    \node[state] (A) at (1,2) {};
    \node[state] (B) at (2,2) {};
    \node[state] (C) at (3,2) {};
    \node[state] (D) at (4,2) {};
    
    \node[thick] at (0,1) {$R$};
    \node[thick] at (0,2) {$P$};
    \node[thick] at (1.5,3) {$I$};
    \node[thick] at (3.5,3) {$E$};
    
    \draw[dotted, thick] (2.5,0.5) -- (2.5,2.5);
    
    \draw[every loop]
    (D) edge[bend right] node [scale=1.2, above=0.1 of C] {$u_2\times 1$} (C)
    (C) edge[bend right] node [scale=1.2, above=0.1 of B] {} (B);

\end{tikzpicture}
\end{minipage}
}

\begin{minipage}{0.24\textwidth}
Condition:
$x_{1}=0$, $x_{2}=0$, $y=0$, $\mathrm{TA}_{4}=\text{E}$.\\
Thus, Type II, $x_2=0$,  \\
$C=1$.  
\end{minipage}
\resizebox{0.18\textwidth}{!}{
\begin{minipage}{0.24\textwidth}
\begin{tikzpicture}[node distance = .35cm, font=\Huge]
    \tikzstyle{every node}=[scale=0.35]
    
    \node[state] (E) at (1,1) {};
    \node[state] (F) at (2,1) {};
    \node[state] (G) at (3,1) {};
    \node[state] (H) at (4,1) {};
  
    \node[state] (A) at (1,2) {};
    \node[state] (B) at (2,2) {};
    \node[state] (C) at (3,2) {};
    \node[state] (D) at (4,2) {};
    
    \node[thick] at (0,1) {$R$};
    \node[thick] at (0,2) {$P$};
    \node[thick] at (1.5,3) {$I$};
    \node[thick] at (3.5,3) {$E$};
    
    \draw[dotted, thick] (2.5,0.5) -- (2.5,2.5);
    
    \draw[every loop]
    (D) edge[bend right] node [scale=1.2, above=0.1 of C] {$u_1\times1$} (C)
    (C) edge[bend right] node [scale=1.2, above=0.1 of B] {} (B);

\end{tikzpicture}
\end{minipage}
}


\begin{minipage}{0.24\textwidth}
Condition:
$x_{1}=1$, $x_{2}=1$, $y=1$, $\mathrm{TA}_{4}=\text{I}$.\\
Thus, Type I, $ x_{2}=1$, \\
$C=0$.  
\end{minipage}
\resizebox{0.18\textwidth}{!}{
\begin{minipage}{0.24\textwidth}
\begin{tikzpicture}[node distance = .35cm, font=\Huge]
    \tikzstyle{every node}=[scale=0.35]
    
    \node[state] (E) at (1,1) {};
    \node[state] (F) at (2,1) {};
    \node[state] (G) at (3,1) {};
    \node[state] (H) at (4,1) {};
  
    \node[state] (A) at (1,2) {};
    \node[state] (B) at (2,2) {};
    \node[state] (C) at (3,2) {};
    \node[state] (D) at (4,2) {};
    
    \node[thick] at (0,1) {$R$};
    \node[thick] at (0,2) {$P$};
    \node[thick] at (1.5,3) {$I$};
    \node[thick] at (3.5,3) {$E$};
    
    \draw[dotted, thick] (2.5,0.5) -- (2.5,2.5);
    
    \draw[every loop]
    (G) edge[bend left] node [scale=1.2, above=0.1 of C] {} (H)
    (H) edge[loop right] node [scale=1.2, below=0.1 of H] {$u_1\frac{1}{s}$} (H);

\end{tikzpicture}
\end{minipage}
}

\begin{minipage}{0.24\textwidth}
Condition:
$x_{1}=1$, $x_{2}=0$, $y=0$, $\mathrm{TA}_{4}=\text{I}$.\\
Thus, Type II, $x_{2}=0$, \\
$C=1$.  
\end{minipage}
\resizebox{0.18\textwidth}{!}{
\begin{minipage}{0.24\textwidth}
\begin{tikzpicture}[node distance = .35cm, font=\Huge]
    \tikzstyle{every node}=[scale=0.35]
    
    \node[state] (E) at (1,1) {};
    \node[state] (F) at (2,1) {};
    \node[state] (G) at (3,1) {};
    \node[state] (H) at (4,1) {};
  
    \node[state] (A) at (1,2) {};
    \node[state] (B) at (2,2) {};
    \node[state] (C) at (3,2) {};
    \node[state] (D) at (4,2) {};
    
    \node[thick] at (0,1) {$R$};
    \node[thick] at (0,2) {$P$};
    \node[thick] at (1.5,3) {$I$};
    \node[thick] at (3.5,3) {$E$};
    
    \draw[dotted, thick] (2.5,0.5) -- (2.5,2.5);
    
    \draw[every loop]
    (D) edge[bend right] node [scale=1.2, above=0.1 of C] {$u_2\times1$} (C)
    (C) edge[bend right] node [scale=1.2, above=0.1 of B] {} (B);

\end{tikzpicture}
\end{minipage}
}

\begin{minipage}{0.24\textwidth}
Condition:
$x_{1}=0$, $x_{2}=0$, $y=0$,  $\mathrm{TA}_{4}=\text{I}$.\\
Thus, Type II, $ x_{2}=0$, \\
$C=1$.  
\end{minipage}
\resizebox{0.18\textwidth}{!}{
\begin{minipage}{0.24\textwidth}
\begin{tikzpicture}[node distance = .35cm, font=\Huge]
    \tikzstyle{every node}=[scale=0.35]
    
    \node[state] (E) at (1,1) {};
    \node[state] (F) at (2,1) {};
    \node[state] (G) at (3,1) {};
    \node[state] (H) at (4,1) {};
  
    \node[state] (A) at (1,2) {};
    \node[state] (B) at (2,2) {};
    \node[state] (C) at (3,2) {};
    \node[state] (D) at (4,2) {};
    
    \node[thick] at (0,1) {$R$};
    \node[thick] at (0,2) {$P$};
    \node[thick] at (1.5,3) {$I$};
    \node[thick] at (3.5,3) {$E$};
    
    \draw[dotted, thick] (2.5,0.5) -- (2.5,2.5);
    
    \draw[every loop]
    (D) edge[bend right] node [scale=1.2, above=0.1 of C] {$u_2\times1$} (C)
    (C) edge[bend right] node [scale=1.2, above=0.1 of B] {} (B);

\end{tikzpicture}
\end{minipage}
}

\subsubsection{Study $\mathrm{TA}_{4}$ with Action \textsl{Include}}

\hspace{0.55cm}\begin{minipage}{0.24\textwidth}
Condition:
$x_{1}=1$, $x_{2}=1$, $y=1$, $\mathrm{TA}_{3}=\text{E}$.\\
Thus, Type I, $ \neg x_{2}=0$, \\
$C=\neg x_{2}=0$.  
\end{minipage}
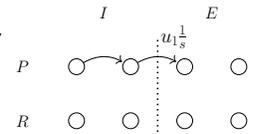
\resizebox{0.18\textwidth}{!}{
\begin{minipage}{0.24\textwidth}
\begin{tikzpicture}[node distance = .35cm, font=\Huge]
    \tikzstyle{every node}=[scale=0.35]
    
    \node[state] (E) at (1,1) {};
    \node[state] (F) at (2,1) {};
    \node[state] (G) at (3,1) {};
    \node[state] (H) at (4,1) {};
  
    \node[state] (A) at (1,2) {};
    \node[state] (B) at (2,2) {};
    \node[state] (C) at (3,2) {};
    \node[state] (D) at (4,2) {};
    
    \node[thick] at (0,1) {$R$};
    \node[thick] at (0,2) {$P$};
    \node[thick] at (1.5,3) {$I$};
    \node[thick] at (3.5,3) {$E$};
    
    \draw[dotted, thick] (2.5,0.5) -- (2.5,2.5);
    
    \draw[every loop]
    (A) edge[bend left] node [scale=1.2, above=0.1 of C] {} (B)
    (B) edge[bend left] node [scale=1.2, above=0.1 of C] {$~~~~~~u_1\frac{1}{s}$} (C);

\end{tikzpicture}
\end{minipage}
}

\begin{minipage}{0.24\textwidth}
Condition:
$x_{1}=1$, $x_{2}=1$, $y=1$, $\mathrm{TA}_{3}=\text{I}$.\\
Thus, Type I, $ \neg x_{2}=0$, \\
$C=\neg x_{2} \wedge x_{2}=0$.  
\end{minipage}
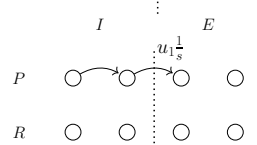
\resizebox{0.18\textwidth}{!}{
\begin{minipage}{0.24\textwidth}
\begin{tikzpicture}[node distance = .35cm, font=\Huge]
    \tikzstyle{every node}=[scale=0.35]
    
    \node[state] (E) at (1,1) {};
    \node[state] (F) at (2,1) {};
    \node[state] (G) at (3,1) {};
    \node[state] (H) at (4,1) {};
  
    \node[state] (A) at (1,2) {};
    \node[state] (B) at (2,2) {};
    \node[state] (C) at (3,2) {};
    \node[state] (D) at (4,2) {};
    
    \node[thick] at (0,1) {$R$};
    \node[thick] at (0,2) {$P$};
    \node[thick] at (1.5,3) {$I$};
    \node[thick] at (3.5,3) {$E$};
    
    \draw[dotted, thick] (2.5,0.5) -- (2.5,2.5);
    
    \draw[every loop]
    (A) edge[bend left] node [scale=1.2, below=0.1 of B] {} (B)
    (B) edge[bend left] node [scale=1.2, above=0.1 of C] {$~~~~~~u_1\frac{1}{s}$} (C);

\end{tikzpicture}
\end{minipage}
}

\subsubsection{Study $\mathrm{TA}_{4}$ with Action \textsl{Exclude}}\label{344}


\hspace{0.55cm}\begin{minipage}{0.24\textwidth}
Condition:
$x_{1}=1$, $x_{2}=1$, $y=1$, $\mathrm{TA}_{3}=\text{E}$.\\
Thus, Type I, $ \neg x_{2}=0$, \\
$C=1$.  
\end{minipage}
\resizebox{0.18\textwidth}{!}{
\begin{minipage}{0.24\textwidth}
\begin{tikzpicture}[node distance = .35cm, font=\Huge]
    \tikzstyle{every node}=[scale=0.35]
    
    \node[state] (E) at (1,1) {};
    \node[state] (F) at (2,1) {};
    \node[state] (G) at (3,1) {};
    \node[state] (H) at (4,1) {};
  
    \node[state] (A) at (1,2) {};
    \node[state] (B) at (2,2) {};
    \node[state] (C) at (3,2) {};
    \node[state] (D) at (4,2) {};
    
    \node[thick] at (0,1) {$R$};
    \node[thick] at (0,2) {$P$};
    \node[thick] at (1.5,3) {$I$};
    \node[thick] at (3.5,3) {$E$};
    
    \draw[dotted, thick] (2.5,0.5) -- (2.5,2.5);
    
    \draw[every loop]
    (H) edge[loop right] node [scale=1.2, below=0.1 of H] {$u_1\frac{1}{s}$} (H)
    (G) edge[bend left] node [scale=1.2, below=0.1 of C] {} (H);

\end{tikzpicture}
\end{minipage}
}

\begin{minipage}{0.24\textwidth}
Condition:
$x_{1}=0$, $x_{2}=1$, $y=0$, $\mathrm{TA}_{3}=\text{E}$.\\
Thus, Type II, $ \neg x_{2}=0$, \\
$C=1$.  
\end{minipage}
\resizebox{0.18\textwidth}{!}{
\begin{minipage}{0.24\textwidth}
\begin{tikzpicture}[node distance = .35cm, font=\Huge]
    \tikzstyle{every node}=[scale=0.35]
    
    \node[state] (E) at (1,1) {};
    \node[state] (F) at (2,1) {};
    \node[state] (G) at (3,1) {};
    \node[state] (H) at (4,1) {};
  
    \node[state] (A) at (1,2) {};
    \node[state] (B) at (2,2) {};
    \node[state] (C) at (3,2) {};
    \node[state] (D) at (4,2) {};
    
    \node[thick] at (0,1) {$R$};
    \node[thick] at (0,2) {$P$};
    \node[thick] at (1.5,3) {$I$};
    \node[thick] at (3.5,3) {$E$};
    
    \draw[dotted, thick] (2.5,0.5) -- (2.5,2.5);
    
    \draw[every loop]
    (D) edge[bend right] node [scale=1.2, above=0.1 of C] {$u_2\times 1$} (C)
    (C) edge[bend right] node [scale=1.2, above=0.1 of B] {} (B);

\end{tikzpicture}
\end{minipage}
}

\begin{minipage}{0.24\textwidth}
Condition:
$x_{1}=1$, $x_{2}=1$, $y=1$, $\mathrm{TA}_{3}=\text{I}$.\\
Thus, Type I, $ \neg x_{2}=0$, \\
$C=1$.  
\end{minipage}
\resizebox{0.18\textwidth}{!}{
\begin{minipage}{0.24\textwidth}
\begin{tikzpicture}[node distance = .35cm, font=\Huge]
    \tikzstyle{every node}=[scale=0.35]
    
    \node[state] (E) at (1,1) {};
    \node[state] (F) at (2,1) {};
    \node[state] (G) at (3,1) {};
    \node[state] (H) at (4,1) {};
  
    \node[state] (A) at (1,2) {};
    \node[state] (B) at (2,2) {};
    \node[state] (C) at (3,2) {};
    \node[state] (D) at (4,2) {};
    
    \node[thick] at (0,1) {$R$};
    \node[thick] at (0,2) {$P$};
    \node[thick] at (1.5,3) {$I$};
    \node[thick] at (3.5,3) {$E$};
    
    \draw[dotted, thick] (2.5,0.5) -- (2.5,2.5);
    
    \draw[every loop]
    (H) edge[loop right] node [scale=1.2, below=0.1 of H] {$u_1\frac{1}{s}$} (H)
    (G) edge[bend left] node [scale=1.2, below=0.1 of C] {} (H);

\end{tikzpicture}
\end{minipage}
}

\begin{minipage}{0.24\textwidth}
Condition:
$x_{1}=0$, $x_{2}=1$, $y=0$, $\mathrm{TA}_{3}=\text{I}$.\\
Thus, Type II, $ \neg x_{2}=0$, \\
$C=1$.  
\end{minipage}
\resizebox{0.18\textwidth}{!}{
\begin{minipage}{0.24\textwidth}
\begin{tikzpicture}[node distance = .35cm, font=\Huge]
    \tikzstyle{every node}=[scale=0.35]
    
    \node[state] (E) at (1,1) {};
    \node[state] (F) at (2,1) {};
    \node[state] (G) at (3,1) {};
    \node[state] (H) at (4,1) {};
  
    \node[state] (A) at (1,2) {};
    \node[state] (B) at (2,2) {};
    \node[state] (C) at (3,2) {};
    \node[state] (D) at (4,2) {};
    
    \node[thick] at (0,1) {$R$};
    \node[thick] at (0,2) {$P$};
    \node[thick] at (1.5,3) {$I$};
    \node[thick] at (3.5,3) {$E$};
    
    \draw[dotted, thick] (2.5,0.5) -- (2.5,2.5);
    
    \draw[every loop]
    (D) edge[bend right] node [scale=1.2, above=0.1 of C] {$u_2\times 1$} (C)
    (C) edge[bend right] node [scale=1.2, above=0.1 of B] {} (B);

\end{tikzpicture}
\end{minipage}
}

\subsection{Case 4: Include Both $\neg x_{1}$ and $x_{1}$}\label{case4}
For Case 4, we assume that the actions of TAs for the first bit are frozen as $\mathrm{TA}_{1}=\text{I}$ and $\mathrm{TA}_{2}=\text{I}$, and thus $C=$~\textbf{0 always}. Similarly, we also have 4 situations, detailed in subsections~\ref{331}-\ref{334}.

\subsubsection{Study $\mathrm{TA}_{3}$ with Action \textsl{Include}}\label{331}

\hspace{0.55cm}\begin{minipage}{0.24\textwidth}
Condition:
$x_{1}=1$, $x_{2}=1$, $y=1$, $\mathrm{TA}_{4}=\text{E}$.\\
Thus, Type I, $x_{2}=1$, \\$C=0$.  
\end{minipage}
\resizebox{0.18\textwidth}{!}{
\begin{minipage}{0.24\textwidth}
\begin{tikzpicture}[node distance = .35cm, font=\Huge]
    \tikzstyle{every node}=[scale=0.35]
    
    \node[state] (E) at (1,1) {};
    \node[state] (F) at (2,1) {};
    \node[state] (G) at (3,1) {};
    \node[state] (H) at (4,1) {};
  
    \node[state] (A) at (1,2) {};
    \node[state] (B) at (2,2) {};
    \node[state] (C) at (3,2) {};
    \node[state] (D) at (4,2) {};
    
    \node[thick] at (0,1) {$R$};
    \node[thick] at (0,2) {$P$};
    \node[thick] at (1.5,3) {$I$};
    \node[thick] at (3.5,3) {$E$};
    
    \draw[dotted, thick] (2.5,0.5) -- (2.5,2.5);
    
    \draw[every loop]
    (A) edge[bend left] node [scale=1.2, above=0.1 of C] {} (B)
    (B) edge[bend left] node [scale=1.2, above=0.1 of C] {$~~~~~u_1\frac{1}{s}$} (C);

\end{tikzpicture}
\end{minipage}
}

\begin{minipage}{0.24\textwidth}
Condition:
$x_{1}=1$, $x_{2}=1$, $y=1$, $\mathrm{TA}_{4}=\text{I}$.\\
Thus, Type I, $x_{2}=1$, \\
$C=0$.  
\end{minipage}
\resizebox{0.18\textwidth}{!}{
\begin{minipage}{0.24\textwidth}
\begin{tikzpicture}[node distance = .35cm, font=\Huge]
    \tikzstyle{every node}=[scale=0.35]
    
    \node[state] (E) at (1,1) {};
    \node[state] (F) at (2,1) {};
    \node[state] (G) at (3,1) {};
    \node[state] (H) at (4,1) {};
  
    \node[state] (A) at (1,2) {};
    \node[state] (B) at (2,2) {};
    \node[state] (C) at (3,2) {};
    \node[state] (D) at (4,2) {};
    
    \node[thick] at (0,1) {$R$};
    \node[thick] at (0,2) {$P$};
    \node[thick] at (1.5,3) {$I$};
    \node[thick] at (3.5,3) {$E$};
    
    \draw[dotted, thick] (2.5,0.5) -- (2.5,2.5);
    
    \draw[every loop]
    (A) edge[bend left] node [scale=1.2, above=0.1 of C] {} (B)
    (B) edge[bend left] node [scale=1.2, above=0.1 of C] {$~~~~~u_1\frac{1}{s}$} (C);

\end{tikzpicture}
\end{minipage}
}

\subsubsection{Study $\mathrm{TA}_{3}$ with Action \textsl{Exclude}}

\hspace{0.55cm}\begin{minipage}{0.24\textwidth}
Condition:
$x_{1}=1$, $x_{2}=1$, $y=1$, $\mathrm{TA}_{4}=\text{E}$.\\
Thus, Type I,  $x_{2}=1$, \\
$C=0$.  
\end{minipage}
\resizebox{0.18\textwidth}{!}{
\begin{minipage}{0.24\textwidth}
\begin{tikzpicture}[node distance = .35cm, font=\Huge]
    \tikzstyle{every node}=[scale=0.35]
    
    \node[state] (E) at (1,1) {};
    \node[state] (F) at (2,1) {};
    \node[state] (G) at (3,1) {};
    \node[state] (H) at (4,1) {};
  
    \node[state] (A) at (1,2) {};
    \node[state] (B) at (2,2) {};
    \node[state] (C) at (3,2) {};
    \node[state] (D) at (4,2) {};
    
    \node[thick] at (0,1) {$R$};
    \node[thick] at (0,2) {$P$};
    \node[thick] at (1.5,3) {$I$};
    \node[thick] at (3.5,3) {$E$};
    
    \draw[dotted, thick] (2.5,0.5) -- (2.5,2.5);
    
    \draw[every loop]
    (G) edge[bend left] node [scale=1.2, above=0.1 of C] {} (H)
    (H) edge[loop right] node [scale=1.2, below=0.1 of H] {$u_1\frac{1}{s}$} (H);

\end{tikzpicture}
\end{minipage}
}

\begin{minipage}{0.24\textwidth}
Condition:
$x_{1}=1$, $x_{2}=1$, $y=1$, $\mathrm{TA}_{4}=\text{I}$.\\
Thus, Type I, $x_{2}=1$, \\
$C=0$.
\end{minipage}
\resizebox{0.18\textwidth}{!}{
\begin{minipage}{0.24\textwidth}
\begin{tikzpicture}[node distance = .35cm, font=\Huge]
    \tikzstyle{every node}=[scale=0.35]
    
    \node[state] (E) at (1,1) {};
    \node[state] (F) at (2,1) {};
    \node[state] (G) at (3,1) {};
    \node[state] (H) at (4,1) {};
  
    \node[state] (A) at (1,2) {};
    \node[state] (B) at (2,2) {};
    \node[state] (C) at (3,2) {};
    \node[state] (D) at (4,2) {};
    
    \node[thick] at (0,1) {$R$};
    \node[thick] at (0,2) {$P$};
    \node[thick] at (1.5,3) {$I$};
    \node[thick] at (3.5,3) {$E$};
    
    \draw[dotted, thick] (2.5,0.5) -- (2.5,2.5);
    
    \draw[every loop]
    (G) edge[bend left] node [scale=1.2, above=0.1 of C] {} (H)
    (H) edge[loop right] node [scale=1.2, below=0.1 of H] {$u_1\frac{1}{s}$} (H);

\end{tikzpicture}
\end{minipage}
}

\subsubsection{Study $\mathrm{TA}_{4}$ with Action \textsl{Include}}

\hspace{0.55cm}\begin{minipage}{0.24\textwidth}
Condition:
$x_{1}=1$, $x_{2}=1$, $y=1$, $\mathrm{TA}_{3}=\text{E}$.\\
Thus, Type I, $\neg x_{2}=0$, \\
$C=0$.  
\end{minipage}
\resizebox{0.18\textwidth}{!}{
\begin{minipage}{0.24\textwidth}
\begin{tikzpicture}[node distance = .35cm, font=\Huge]
    \tikzstyle{every node}=[scale=0.35]
    
    \node[state] (E) at (1,1) {};
    \node[state] (F) at (2,1) {};
    \node[state] (G) at (3,1) {};
    \node[state] (H) at (4,1) {};
  
    \node[state] (A) at (1,2) {};
    \node[state] (B) at (2,2) {};
    \node[state] (C) at (3,2) {};
    \node[state] (D) at (4,2) {};
    
    \node[thick] at (0,1) {$R$};
    \node[thick] at (0,2) {$P$};
    \node[thick] at (1.5,3) {$I$};
    \node[thick] at (3.5,3) {$E$};
    
    \draw[dotted, thick] (2.5,0.5) -- (2.5,2.5);
    
    \draw[every loop]
    (A) edge[bend left] node [scale=1.2, above=0.1 of C] {} (B)
    (B) edge[bend left] node [scale=1.2, above=0.1 of C] {$~~~~u_1\frac{1}{s}$} (C);

\end{tikzpicture}
\end{minipage}
}

\begin{minipage}{0.24\textwidth}
Condition:
$x_{1}=1$, $x_{2}=1$, $y=1$, $\mathrm{TA}_{3}=\text{I}$.\\
Thus, Type I, $\neg x_{2}=0$, \\
$C=0$.  
\end{minipage}
\resizebox{0.18\textwidth}{!}{
\begin{minipage}{0.24\textwidth}
\begin{tikzpicture}[node distance = .35cm, font=\Huge]
    \tikzstyle{every node}=[scale=0.35]
    
    \node[state] (E) at (1,1) {};
    \node[state] (F) at (2,1) {};
    \node[state] (G) at (3,1) {};
    \node[state] (H) at (4,1) {};
  
    \node[state] (A) at (1,2) {};
    \node[state] (B) at (2,2) {};
    \node[state] (C) at (3,2) {};
    \node[state] (D) at (4,2) {};
    
    \node[thick] at (0,1) {$R$};
    \node[thick] at (0,2) {$P$};
    \node[thick] at (1.5,3) {$I$};
    \node[thick] at (3.5,3) {$E$};
    
    \draw[dotted, thick] (2.5,0.5) -- (2.5,2.5);
    
    \draw[every loop]
    (A) edge[bend left] node [scale=1.2, above=0.1 of C] {} (B)
    (B) edge[bend left] node [scale=1.2, above=0.1 of C] {$~~~~~u_1\frac{1}{s}$} (C);

\end{tikzpicture}
\end{minipage}
}

\subsubsection{Study $\mathrm{TA}_{4}$ with Action Exclude}\label{334}

\hspace{0.55cm}\begin{minipage}{0.24\textwidth}
Condition:
$x_{1}=1$, $x_{2}=1$, $y=1$,  $\mathrm{TA}_{3}=\text{E}$.\\
Thus, Type I, $\neg x_{2}=0$, \\
$C=0$.
\end{minipage}
\resizebox{0.18\textwidth}{!}{
\begin{minipage}{0.24\textwidth}
\begin{tikzpicture}[node distance = .35cm, font=\Huge]
    \tikzstyle{every node}=[scale=0.35]
    
    \node[state] (E) at (1,1) {};
    \node[state] (F) at (2,1) {};
    \node[state] (G) at (3,1) {};
    \node[state] (H) at (4,1) {};
  
    \node[state] (A) at (1,2) {};
    \node[state] (B) at (2,2) {};
    \node[state] (C) at (3,2) {};
    \node[state] (D) at (4,2) {};
    
    \node[thick] at (0,1) {$R$};
    \node[thick] at (0,2) {$P$};
    \node[thick] at (1.5,3) {$I$};
    \node[thick] at (3.5,3) {$E$};
    
    \draw[dotted, thick] (2.5,0.5) -- (2.5,2.5);
    
    \draw[every loop]
    (G) edge[bend left] node [scale=1.2, above=0.1 of C] {} (H)
    (H) edge[loop right] node [scale=1.2, below=0.1 of H] {$u_1\frac{1}{s}$} (H);

\end{tikzpicture}
\end{minipage}
}

\begin{minipage}{0.24\textwidth}
Condition:
$x_{1}=1$, $x_{2}=1$, $y=1$, $\mathrm{TA}_{3}=\text{I}$.\\
Thus, Type I, $\neg x_{2}=0$, \\
$C=0$.
\end{minipage}
\resizebox{0.18\textwidth}{!}{
\begin{minipage}{0.24\textwidth}
\begin{tikzpicture}[node distance = .35cm, font=\Huge]
    \tikzstyle{every node}=[scale=0.35]
    
    \node[state] (E) at (1,1) {};
    \node[state] (F) at (2,1) {};
    \node[state] (G) at (3,1) {};
    \node[state] (H) at (4,1) {};
  
    \node[state] (A) at (1,2) {};
    \node[state] (B) at (2,2) {};
    \node[state] (C) at (3,2) {};
    \node[state] (D) at (4,2) {};
    
    \node[thick] at (0,1) {$R$};
    \node[thick] at (0,2) {$P$};
    \node[thick] at (1.5,3) {$I$};
    \node[thick] at (3.5,3) {$E$};
    
    \draw[dotted, thick] (2.5,0.5) -- (2.5,2.5);
    
    \draw[every loop]
    (G) edge[bend left] node [scale=1.2, above=0.1 of C] {} (H)
    (H) edge[loop right] node [scale=1.2, below=0.1 of H] {$u_1\frac{1}{s}$} (H);

\end{tikzpicture}
\end{minipage}
}

So far, we have gone through, exhaustively, the transitions of $\mathrm{TA}_{3}$ and $\mathrm{TA}_{4}$ for all possible training samples and system states. In what follows, we will summarize the direction of transitions and study the convergence properties of the system for the given training samples.


\subsection{Summarize of the Directions of Transitions in Different Cases}\label{summary}
Based on the analysis above, we summarize here what happens to $\mathrm{TA}_{3}$ and $\mathrm{TA}_{4}$,  given different status (Cases) of $\mathrm{TA}_{1}$ and $\mathrm{TA}_{2}$.  More specifically, we will summarize here the directions of the transitions for the TAs. For example, ``$\mathrm{TA}_{3} \Rightarrow$ E" means that $\mathrm{TA}_{3}$ will move towards the action ``Exclude", while ``$\mathrm{TA}_{4} \Rightarrow$ E or I" means $\mathrm{TA}_{4}$ transits towards either ``Exclude" or ``Include". 

\textbf{Scenario 1:} Study $\mathrm{TA}_{3} =\text{I}$ and $\mathrm{TA}_{4} =\text{I}$.

\begin{minipage}{0.225\textwidth}
\textbf{Case 1}, we have: \\
$\mathrm{TA}_{3} \Rightarrow$ E.\\
$\mathrm{TA}_{4} \Rightarrow$ E.\\
\textbf{Case 2}, we have:\\
$\mathrm{TA}_{3} \Rightarrow$ E.\\
$\mathrm{TA}_{4} \Rightarrow$ E.
\end{minipage}
\begin{minipage}{0.225\textwidth}
\textbf{Case 3}, we have: \\
$\mathrm{TA}_{3} \Rightarrow$ E.\\
$\mathrm{TA}_{4} \Rightarrow$ E.\\
\textbf{Case 4}, we have: \\
$\mathrm{TA}_{3} \Rightarrow$ E.\\
$\mathrm{TA}_{4} \Rightarrow$ E.
\end{minipage}

\vspace{.5cm}

From the facts presented above, we can confirm that regardless the state of $\mathrm{TA}_{1}$ and $\mathrm{TA}_{2}$,
if $\mathrm{TA}_{3}$ = I and $\mathrm{TA}_{4}$ = I,
they ($\mathrm{TA}_{3}$ and $\mathrm{TA}_{4}$) will eventually move out of their states.

\vspace{0.25cm}

\textbf{Scenario 2:} Study $\mathrm{TA}_{3} =\text{I}$ and $\mathrm{TA}_{4} =\text{E}$.

\begin{minipage}{0.225\textwidth}
\textbf{Case 1}, we have: \\
$\mathrm{TA}_{3} \Rightarrow$ E.\\
$\mathrm{TA}_{4} \Rightarrow$ E or I.\\
\textbf{Case 2}, we have:\\
$\mathrm{TA}_{3} \Rightarrow$ I.\\
$\mathrm{TA}_{4} \Rightarrow$ E.
\end{minipage}
\begin{minipage}{.225\textwidth}
\textbf{Case 3}, we have: \\
$\mathrm{TA}_{3} \Rightarrow$ I.\\
$\mathrm{TA}_{4} \Rightarrow$ E or I.\\
\textbf{Case 4}, we have: \\
$\mathrm{TA}_{3} \Rightarrow$ E.\\
$\mathrm{TA}_{4} \Rightarrow$ E.
\end{minipage}

\vspace{0.25cm}

For Scenario 2 Case 2, we can observe that if $\mathrm{TA}_{3}=\text{I}$, $\mathrm{TA}_{4}=\text{E}$, $\mathrm{TA}_{1}=\text{I}$, and $\mathrm{TA}_{2}=\text{E}$, $\mathrm{TA}_{3}$ will move deeper to ``include" and $\mathrm{TA}_{4}$ will go deeper to ``exclude". It is not difficult to derive also that $\mathrm{TA}_{1}$ will move deeper to ``include" and $\mathrm{TA}_{2}$ will transfer deeper to ``exclude" in this circumstance. This tells us that the TAs in states $\mathrm{TA}_{3}=\text{I}$, $\mathrm{TA}_{4}=\text{E}$, $\mathrm{TA}_{1}=\text{I}$, and $\mathrm{TA}_{2}=\text{E}$, reinforce each other to move deeper to their corresponding directions and they therefore construct an absorbing state of the system. If it is the only absorbing state, we can conclude that the TM converge to the intended ``AND" operation. 

In Scenario 2, we can observe for Cases 1, 3, and 4, the actions for $\mathrm{TA}_{3}$ and $\mathrm{TA}_{4}$ are not absorbing because the TAs will not be reinforced to move monotonically deeper to the states of the corresponding actions for difference cases.

For Scenario 2, Case 3, $\mathrm{TA}_{4}$ has two possible directions to transit, I or E, depending on the input of the training sample. For action exclude, it will be reinforced when training sample $x_1=1$ and $x_2=1$ is given, based on Type I feedback. However, $\mathrm{TA}_{4}$ will transit towards ``include" side when training sample $x_1=0$ and $x_2=1$ is given, due to Type II feedback. Therefore, the direction of the transition for $\mathrm{TA}_{4}$ is I or E, depending on the training samples. In the following paragraphs, when ``or" appears in the transition direction, the same concept applies.

\vspace{0.25cm}

\textbf{Scenario 3:} Study $\mathrm{TA}_{3} =\text{E}$ and $\mathrm{TA}_{4} =\text{I}$.

\begin{minipage}{.225\textwidth}
\textbf{Case 1}, we have: \\
$\mathrm{TA}_{3} \Rightarrow$ E or I.\\
$\mathrm{TA}_{4} \Rightarrow$ E.\\
\textbf{Case 2}, we can see that:\\
$\mathrm{TA}_{3} \Rightarrow$ E or I.\\
$\mathrm{TA}_{4} \Rightarrow$ E.
\end{minipage}
\begin{minipage}{.225\textwidth}
\textbf{Case 3}, we have: \\
$\mathrm{TA}_{3} \Rightarrow$ E or I.\\
$\mathrm{TA}_{4} \Rightarrow$ E.\\
\textbf{Case 4}, we have: \\
$\mathrm{TA}_{3} \Rightarrow$ E.\\
$\mathrm{TA}_{4} \Rightarrow$ E.
\end{minipage}

\vspace{0.25cm}
In Scenario 3, we can see that the actions for $\mathrm{TA}_{3}=\text{E}$ and $\mathrm{TA}_{4} =\text{I}$ are not absorbing because the TAs will not be reinforced to move deeper to the states of the corresponding actions.
\vspace{0.25cm}

\textbf{Scenario 4:} Study $\mathrm{TA}_{3} =\text{E}$ and $\mathrm{TA}_{4} =\text{E}$.

\begin{minipage}{.225\textwidth}
\textbf{Case 1}, we have: \\
$\mathrm{TA}_{3} \Rightarrow$ I or E.\\
$\mathrm{TA}_{4} \Rightarrow$ I or E.\\
\textbf{Case 2}, we have:\\
$\mathrm{TA}_{3} \Rightarrow$ I.\\
$\mathrm{TA}_{4} \Rightarrow$ E.
\end{minipage}
\begin{minipage}{.225\textwidth}
\textbf{Case 3}, we have: \\
$\mathrm{TA}_{3} \Rightarrow$ I.\\
$\mathrm{TA}_{4} \Rightarrow$ I or E.\\
\textbf{Case 4}, we have: \\
$\mathrm{TA}_{3} \Rightarrow$ E.\\
$\mathrm{TA}_{4} \Rightarrow$ E.
\end{minipage}

\vspace{0.25cm}
In Scenario 4, we see that, the actions for $\mathrm{TA}_{3}=\text{E}$ and $\mathrm{TA}_{4} =\text{E}$ seem to be an absorbing state, because the states of TAs will move deeper in Case 4. After a revisit of the condition for Case 4, i.e.,  include both $\neg x_{1}$ and $x_{1}$, we understand that this condition is not absorbing. In fact, when $\mathrm{TA}_{1}$ and $\mathrm{TA}_{2}$ both have ``Include" as their actions, they monotonically move towards ``Exclude". Therefore, from the overall system's perspective, the system state $\mathrm{TA}_{1}=\text{I}$, $\mathrm{TA}_{2}=\text{I}$, $\mathrm{TA}_{3}=\text{E}$, and $\mathrm{TA}_{4}=\text{E}$ is not absorbing. For the other cases in the this scenario, there is no absorbing state. 

Based on the above analysis, we understand that there is only one absorbing condition in the system, namely,  $\mathrm{TA}_{1}=\text{I}$, $\mathrm{TA}_{2}=\text{E}$, $\mathrm{TA}_{3}=\text{I}$, and $\mathrm{TA}_{4}=\text{E}$,  for the given training samples with AND logic. The same conclusion applies when we freeze the transition of the two TAs for the second bit of the input and study behavior of the first bit of input. Therefore, we can conclude that the TM with only one clause can learn to be the intended AND operator, almost surely, in infinite time horizon.

\section{Convergence of Analysis of the OR Operator}\label{Sec:OR}
Once the convergence of the AND operator is proven, it is self-evident that the convergence of the operator OR is also proven. Clearly, if there is one additional NOT gate that is added after the AND operator, the OR operator is achieved. However, this approach requires a change in the structure of the vanilla TM. In fact, the TM can almost surely converge to the intended OR operator if more clauses are given, and the proof is given presently.

To analyse the OR operator, we assume the training samples with the probability below are given.
\begin{align}\label{orlogic}
P\left ( y=1 | x_{1}=1, x_{2}=1 \right ) = 1, \\
P\left ( y=1 | x_{1}=0, x_{2}=1 \right ) = 1,\nonumber\\
P\left ( y=1 | x_{1}=1, x_{2}=0 \right ) = 1,\nonumber\\
P\left ( y=0 | x_{1}=0, x_{2}=0 \right ) = 1.\nonumber
\end{align}

Clearly, there are three sub-patterns of $x_1$ and $x_2$ that will give $y=1$, i.e., ($x_1=1,~x_2=1$), ($x_1=1,~x_2=0$), and ($x_1=0,~x_2=1$). More specifically, Eq.~(\ref{orlogic}) can be split into three cases,  corresponding to the three sub-patterns:
\begin{align}\label{ANDlogic}
P\left ( y=1 | x_{1}=1, x_{2}=1 \right ) = 1, \\
P\left ( y=0 | x_{1}=0, x_{2}=0 \right ) = 1, \nonumber
\end{align}
\begin{align}\label{orlogic1}
P\left ( y=1 | x_{1}=0, x_{2}=1 \right ) = 1,\\
P\left ( y=0 | x_{1}=0, x_{2}=0 \right ) = 1, \nonumber
\end{align}
and
\begin{align}\label{orlogic2}
P\left ( y=1 | x_{1}=1, x_{2}=0 \right ) = 1,\\
P\left ( y=0 | x_{1}=0, x_{2}=0 \right ) = 1.\nonumber
\end{align}

In what follows, we will first show that the clauses are able to learn each individual sub-pattern shown in Eqs.~(\ref{ANDlogic})-(\ref{orlogic2}) and then we will show the system behavior when more sub-patterns jointly appear in the training samples without~$T$. Thereafter, we conclude the convergence of the system when~$T$ is enabled in the learning. 

\subsection{Ability to Learn All the Sub-patterns of OR Operator}

The convergence analyses of the above three sub-patterns can be derived by reusing the analyses of the sub-patterns of the XOR operator plus the AND operator. For Eq.~(\ref{ANDlogic}), we can see that the TAs will converge to $\mathrm{TA}_{1}=\text{I}$, $\mathrm{TA}_{2}=\text{E}$, $\mathrm{TA}_{3}=\text{I}$, and $\mathrm{TA}_{4}=\text{E}$, by studying the transition diagrams in Subsections~\ref{case1}-\ref{case4} when input samples of $x_1=0$, $x_2=1$ and $x_1=1$, $x_2=0$ are removed.  In more details, the directions of the transitions for different scenarios are summarized below. The 4 cases mentioned below, i.e., \textbf{Case~1--Case 4}, are defined in Subsections~\ref{case1}-\ref{case4}.


{\color{black} 
\textbf{Scenario 1:} Study $\mathrm{TA}_{3} =\text{I}$ and $\mathrm{TA}_{4} =\text{I}$.

\begin{minipage}{0.225\textwidth}
\textbf{Case 1}, we have: \\
$\mathrm{TA}_{3} \Rightarrow$ E.\\
$\mathrm{TA}_{4} \Rightarrow$ E.\\
\textbf{Case 2}, we can see that:\\
$\mathrm{TA}_{3} \Rightarrow$ E.\\
$\mathrm{TA}_{4} \Rightarrow$ E.
\end{minipage}
\begin{minipage}{0.225\textwidth}
\textbf{Case 3}, we have: \\
$\mathrm{TA}_{3} \Rightarrow$ E.\\
$\mathrm{TA}_{4} \Rightarrow$ E.\\
\textbf{Case 4}, we have: \\
$\mathrm{TA}_{3} \Rightarrow$ E.\\
$\mathrm{TA}_{4} \Rightarrow$ E.
\end{minipage}

\vspace{0.5cm}

\textbf{Scenario 2:} Study $\mathrm{TA}_{3} =\text{I}$ and $\mathrm{TA}_{4} =\text{E}$.

\begin{minipage}{.225\textwidth}
\textbf{Case 1}, we have: \\
$\mathrm{TA}_{3} \Rightarrow$ E.\\
$\mathrm{TA}_{4} \Rightarrow$ E.\\
\textbf{Case 2}, we have:\\
$\mathrm{TA}_{3} \Rightarrow$ I.\\
$\mathrm{TA}_{4} \Rightarrow$ E.\\
\end{minipage}
\begin{minipage}{.225\textwidth}
\textbf{Case 3}, we have: \\
$\mathrm{TA}_{3} \Rightarrow$ I.\\
$\mathrm{TA}_{4} \Rightarrow$ E.\\
\textbf{Case 4}, we have: \\
$\mathrm{TA}_{3} \Rightarrow$ E.\\
$\mathrm{TA}_{4} \Rightarrow$ E.

\end{minipage}

\vspace{.25cm}

\textbf{Scenario 3:} Study $\mathrm{TA}_{3} =\text{E}$ and $\mathrm{TA}_{4} =\text{I}$.

\begin{minipage}{.225\textwidth}
\textbf{Case 1}, we have: \\
$\mathrm{TA}_{3} \Rightarrow$ E or I.\\
$\mathrm{TA}_{4} \Rightarrow$ E.\\
\textbf{Case 2}, we have:\\
$\mathrm{TA}_{3} \Rightarrow$ E.\\
$\mathrm{TA}_{4} \Rightarrow$ E.
\end{minipage}
\begin{minipage}{.225\textwidth}
\textbf{Case 3}, we have: \\
$\mathrm{TA}_{3} \Rightarrow$ E or I.\\
$\mathrm{TA}_{4} \Rightarrow$ E.\\
\textbf{Case 4}, we have: \\
$\mathrm{TA}_{3} \Rightarrow$ E.\\
$\mathrm{TA}_{4} \Rightarrow$ E.\\
\end{minipage}

\vspace{.25cm}

\textbf{Scenario 4:} Study $\mathrm{TA}_{3} =\text{E}$ and $\mathrm{TA}_{4} =\text{E}$.

\begin{minipage}{.225\textwidth}
\textbf{Case 1}, we have: \\
$\mathrm{TA}_{3} \Rightarrow$ I or E.\\
$\mathrm{TA}_{4} \Rightarrow$ E.\\
\textbf{Case 2}, we have:\\
$\mathrm{TA}_{3} \Rightarrow$ I.\\
$\mathrm{TA}_{4} \Rightarrow$ E.
\end{minipage}
\begin{minipage}{.225\textwidth}
\textbf{Case 3}, we have: \\
$\mathrm{TA}_{3} \Rightarrow$ I.\\
$\mathrm{TA}_{4} \Rightarrow$ E.\\
\textbf{Case 4}, we have: \\
$\mathrm{TA}_{3} \Rightarrow$ E.\\
$\mathrm{TA}_{4} \Rightarrow$ E.
\end{minipage}

\vspace{0.25cm}
Comparing the analysis with the one in Subsection~\ref{summary}, there is apparently another possible absorbing case, which can be observed in Scenario 2, Case 3, where $\mathrm{TA}_{3}=\text{I}$ and $\mathrm{TA}_{4}=\text{E}$, given $\mathrm{TA}_{1}=\text{E}$ and $\mathrm{TA}_{2}=\text{E}$. However, given $\mathrm{TA}_{3}=\text{I}$ and $\mathrm{TA}_{4}=\text{E}$, the TAs for the first bit, i.e., $\mathrm{TA}_{1}=\text{E}$ and $\mathrm{TA}_{2}=\text{E}$, will not move only towards Exclude. Therefore, they do not reinforce each other to move to deeper states for their current actions. For this reason,  $\mathrm{TA}_{3}=\text{I}$ and $\mathrm{TA}_{4}=\text{E}$, $\mathrm{TA}_{1}=\text{E}$ and $\mathrm{TA}_{2}=\text{E}$, is not an absorbing state.  In addition, given  $\mathrm{TA}_{3}=\text{I}$, $\mathrm{TA}_{4}=\text{E}$,   $\mathrm{TA}_{1}$ and  $\mathrm{TA}_{2}$ with actions E and E will transit towards I and E, encouraging the overall system to move towards I, E, I, and E. Consequently, the system state with $\mathrm{TA}_{1}=\text{I}$, $\mathrm{TA}_{2}=\text{E}$, $\mathrm{TA}_{3}=\text{I}$, and $\mathrm{TA}_{4}=\text{E}$ is still the only absorbing case for the given training samples following Eq.~(\ref{ANDlogic}). 
}


For Eq.~(\ref{orlogic1}), similar to the proof of in Lemma 1 in~\cite{jiao2021convergence}, we can derive that the TAs will converge in $\mathrm{TA}_{1}=\text{E}$, $\mathrm{TA}_{2}=\text{I}$, $\mathrm{TA}_{3}=\text{I}$, and $\mathrm{TA}_{4}=\text{E}$. The transition diagrams for the samples of Eq.~(\ref{orlogic1}) are in fact a subset of the ones presented in Subsection 3.2.1 and Appendix 2 of~\cite{jiao2021convergence}, when the input samples of $x_1=1$ and $x_2=1$ are removed. We summarize below only the directions of transitions. 

The directions of the transitions of the TAs for the second input bit, i.e., $x_2$/$\neg x_2$, when the TAs for the first input bit are frozen, are summarized as follows (based on the subset of the transition diagrams in Subsection 3.2.1 of~\cite{jiao2021convergence}). 

\textbf{Scenario 1:} Study $\mathrm{TA}_3$ = I and $\mathrm{TA}_4$ = I.

\begin{minipage}{0.225\textwidth}
\textbf{Case 1:} we can see that \\
$\mathrm{TA}_3$ $\rightarrow$ E \\
$\mathrm{TA}_4$ $\rightarrow$ E \\
\textbf{Case 2:} we can see that \\
$\mathrm{TA}_3$ $\rightarrow$ E \\
$\mathrm{TA}_4$ $\rightarrow$ E 
\end{minipage}
\begin{minipage}{0.225\textwidth}
\textbf{Case 3:} we can see that \\
$\mathrm{TA}_3$ $\rightarrow$ E \\
$\mathrm{TA}_4$ $\rightarrow$ E \\
\textbf{Case 4:} we can see that \\
$\mathrm{TA}_3$ $\rightarrow$ E \\
$\mathrm{TA}_4$ $\rightarrow$ E 
\end{minipage}

 \vspace{.5cm}

\textbf{Scenario 2:} Study $\mathrm{TA}_3$ = I and $\mathrm{TA}_4=\text{E}$.

\begin{minipage}{0.225\textwidth}
\textbf{Case 1:} we can see that \\
$\mathrm{TA}_3$ $\rightarrow$ I \\
$\mathrm{TA}_4$ $\rightarrow$ E \\
\textbf{Case 2:} we can see that \\
$\mathrm{TA}_3$ $\rightarrow$ E \\
$\mathrm{TA}_4$ $\rightarrow$  E 
\end{minipage}
\begin{minipage}{0.225\textwidth}
{\color{black}\textbf{Case 3:} we can see that \\
$\mathrm{TA}_3$ $\rightarrow$ I \\
$\mathrm{TA}_4$ $\rightarrow$  E }\\
\textbf{Case 4:} we can see that \\
$\mathrm{TA}_3$ $\rightarrow$ E \\
$\mathrm{TA}_4$ $\rightarrow$ E 
\end{minipage}

 \vspace{.5cm}
\textbf{Scenario 3:} Study $\mathrm{TA}_3$ = E and $\mathrm{TA}_4$ = I.

\begin{minipage}{0.225\textwidth}
\textbf{Case 1:} we can see that \\
$\mathrm{TA}_3$ $\rightarrow$ I, or E \\
$\mathrm{TA}_4$ $\rightarrow$ E \\
\textbf{Case 2:} we can see that \\
$\mathrm{TA}_3$ $\rightarrow$ E \\
$\mathrm{TA}_4$ $\rightarrow$ E 
\end{minipage}
\begin{minipage}{0.225\textwidth}
\textbf{Case 3:} we can see that \\
$\mathrm{TA}_3$ $\rightarrow$ I, or E \\
$\mathrm{TA}_4$ $\rightarrow$ E \\
\textbf{Case 4:} we can see that \\
$\mathrm{TA}_3$ $\rightarrow$ E \\
$\mathrm{TA}_4$ $\rightarrow$ E 
\end{minipage}

\vspace{.5cm}

\textbf{Scenario 4:} Study $\mathrm{TA}_3$ = E and $\mathrm{TA}_4$ = E.

\begin{minipage}{0.225\textwidth}
\textbf{Case 1:} we can see that \\
$\mathrm{TA}_3$ $\rightarrow$ I \\
$\mathrm{TA}_4$ $\rightarrow$ E \\
{\color{black}\textbf{Case 2:} we can see that \\
$\mathrm{TA}_3$ $\rightarrow$ E \\
$\mathrm{TA}_4$ $\rightarrow$  E}
\end{minipage}
\begin{minipage}{0.225\textwidth}
\textbf{Case 3:} we can see that \\
$\mathrm{TA}_3$ $\rightarrow$I \\
$\mathrm{TA}_4$ $\rightarrow$ E\\
\textbf{Case 4:} we can see that \\
$\mathrm{TA}_3$ $\rightarrow$ E \\
$\mathrm{TA}_4$ $\rightarrow$ E 
\end{minipage}

\vspace{.5cm}
The directions of the transitions of the TAs for the first input bit, i.e., $x_1$/$\neg x_1$, when the TAs for the second input bit are frozen, are summarized as follows (based on the subset of the transition diagrams in Appendix 2 of~\cite{jiao2021convergence}).

\textbf{Scenario 1:} Study $\mathrm{TA}_1$ = I and $\mathrm{TA}_2$ = I.

\begin{minipage}{0.225\textwidth}
\textbf{Case 1:} we can see that \\
$\mathrm{TA}_1$ $\rightarrow$ E \\
$\mathrm{TA}_2$ $\rightarrow$ E \\
\textbf{Case 2:} we can see that \\
$\mathrm{TA}_1$ $\rightarrow$ E \\
$\mathrm{TA}_2$ $\rightarrow$ E 
\end{minipage}
\begin{minipage}{0.225\textwidth}
\textbf{Case 3:} we can see that \\
$\mathrm{TA}_1$ $\rightarrow$ E \\
$\mathrm{TA}_2$ $\rightarrow$ E \\
\textbf{Case 4:} we can see that \\
$\mathrm{TA}_1$ $\rightarrow$ E \\
$\mathrm{TA}_2$ $\rightarrow$ E 
\end{minipage}

\vspace{.5cm}

\textbf{Scenario 2:} Study $\mathrm{TA}_1$ = I and $\mathrm{TA}_2$ = E.

\begin{minipage}{0.225\textwidth}
\textbf{Case 1:} we can see that \\
$\mathrm{TA}_1$ $\rightarrow$ E \\
$\mathrm{TA}_2$ $\rightarrow$ E \\
\textbf{Case 2:} we can see that \\
$\mathrm{TA}_1$ $\rightarrow$ E \\
$\mathrm{TA}_2$ $\rightarrow$  E 
\end{minipage}
\begin{minipage}{0.225\textwidth}
\textbf{Case 3:} we can see that \\
$\mathrm{TA}_1$ $\rightarrow$ E \\
$\mathrm{TA}_2$ $\rightarrow$ E\\
\textbf{Case 4:} we can see that \\
$\mathrm{TA}_1$ $\rightarrow$ E \\
$\mathrm{TA}_2$ $\rightarrow$ E 
\end{minipage}

\vspace{.5cm}

\textbf{Scenario 3:} Study $\mathrm{TA}_1$ = E and $\mathrm{TA}_2$ = I.

\begin{minipage}{0.225\textwidth}
\textbf{Case 1:} we can see that \\
$\mathrm{TA}_1$ $\rightarrow$ I, or E \\
$\mathrm{TA}_2$ $\rightarrow$ E\\
\textbf{Case 2:} we can see that \\
$\mathrm{TA}_1$ $\rightarrow$ E \\
$\mathrm{TA}_2$ $\rightarrow$ I
\end{minipage}
\begin{minipage}{0.225\textwidth}
\textbf{Case 3:} we can see that \\
$\mathrm{TA}_1$ $\rightarrow$ I \\
$\mathrm{TA}_2$ $\rightarrow$ I\\
\textbf{Case 4:} we can see that \\
$\mathrm{TA}_1$ $\rightarrow$ E \\
$\mathrm{TA}_2$ $\rightarrow$ E 
\end{minipage}

\vspace{.5cm}

\textbf{Scenario 4:} Study $\mathrm{TA}_1$ = E and $\mathrm{TA}_2$ = E.

\begin{minipage}{0.225\textwidth}
\textbf{Case 1:} we can see that \\
$\mathrm{TA}_1$ $\rightarrow$ I, or E \\
$\mathrm{TA}_2$ $\rightarrow$ E\\
\textbf{Case 2:} we can see that \\
$\mathrm{TA}_1$ $\rightarrow$ E\\
$\mathrm{TA}_2$ $\rightarrow$ I 
\end{minipage}
\begin{minipage}{0.225\textwidth}
\textbf{Case 3:} we can see that \\
{\color{black}$\mathrm{TA}_1$ $\rightarrow$ E\\
$\mathrm{TA}_2$ $\rightarrow$  E }\\
\textbf{Case 4:} we can see that \\
$\mathrm{TA}_1$ $\rightarrow$ E \\
$\mathrm{TA}_2$ $\rightarrow$ E 
\end{minipage}
\vspace{.5cm}

By analyzing the transitions of TAs for the two input bits with samples following Eq.~(\ref{orlogic1}), we can conclude that $\mathrm{TA}_{1}=\text{E}$, $\mathrm{TA}_{2}=\text{I}$, $\mathrm{TA}_{3}=\text{I}$, and $\mathrm{TA}_{4}=\text{E}$ is an absorbing state, as the actions of $\mathrm{TA}_{1}$-$\mathrm{TA}_{4}$ reinforce each other to transit to deeper states for the current actions upon various input samples. There are still a few other cases in different scenarios that also seem to be absorbing. But the conditions for those absorbing-like states are not absorbing, making those states not absorbing from system's point of view. For example, the status $\mathrm{TA}_{3}=\text{I}$ and $\mathrm{TA}_{4}=\text{E}$ seems also absorbing in Scenario 2, Case 3, i.e., when $\mathrm{TA}_{1}=\text{E}$ and $\mathrm{TA}_{2}=\text{E}$ hold. However, to make $\mathrm{TA}_{1}=\text{E}$ and $\mathrm{TA}_{2}=\text{E}$ absorbing,  the condition is  $\mathrm{TA}_{3}=\text{I}$ and $\mathrm{TA}_{4}=\text{I}$, or $\mathrm{TA}_{3}=\text{E}$ and $\mathrm{TA}_{4}=\text{E}$. Clearly, the status $\mathrm{TA}_{3}=\text{I}$ and $\mathrm{TA}_{4}=\text{I}$ is not absorbing. For $\mathrm{TA}_{3}=\text{E}$ and $\mathrm{TA}_{4}=\text{E}$ to be absorbing, it is required to have $\mathrm{TA}_{1}=\text{I}$ and $\mathrm{TA}_{2}=\text{I}$ to be absorbing, or  $\mathrm{TA}_{1}=\text{I}$ and $\mathrm{TA}_{2}=\text{E}$ to be absorbing, which are not true. Therefore, all those absorbing-like states are not absorbing. In fact, when $\mathrm{TA}_{3}=\text{I}$,  $\mathrm{TA}_{4}=\text{E}$,  $\mathrm{TA}_{1}=\text{E}$, and $\mathrm{TA}_{2}=\text{E}$ hold, the condition $\mathrm{TA}_{3}=\text{I}$,  $\mathrm{TA}_{4}=\text{E}$ will reinforce $\mathrm{TA}_{1}$ and $\mathrm{TA}_{2}$ to move towards E, I, which is the absorbing state of the system.

Following the same principle, the TAs will converge to $\mathrm{TA}_{1}=\text{I}$, $\mathrm{TA}_{2}=\text{E}$, $\mathrm{TA}_{3}=\text{E}$, and $\mathrm{TA}_{4}=\text{I}$ when training samples from Eq.~(\ref{orlogic2}) are given, according to the proof of Lemma 2 in~\cite{jiao2021convergence}.

\subsection{System Property without $T$}

So far, we show that the clauses is able to converge to the intended operator if the training samples for an individual sub-pattern are given. In what follows, we will show that the system becomes recurrent if any sub-patterns of two training samples are given when $u_1>0$ and $u_2>0$. Specifically, we show that there is no absorbing state for
\begin{align}\label{ANDlogicandxor2}
P\left ( y=1 | x_{1}=1, x_{2}=1 \right ) = 1, \\
P\left ( y=1 | x_{1}=1, x_{2}=0 \right ) = 1, \nonumber\\
P\left ( y=0 | x_{1}=0, x_{2}=0 \right ) = 1, \nonumber
\end{align}

\begin{align}\label{ANDlogicandxor1}
P\left ( y=1 | x_{1}=1, x_{2}=1 \right ) = 1,\\
P\left ( y=1 | x_{1}=0, x_{2}=1 \right ) = 1, \nonumber\\
P\left ( y=0 | x_{1}=0, x_{2}=0 \right ) = 1, \nonumber
\end{align}
and
\begin{align}\label{xor12}
P\left ( y=1 | x_{1}=1, x_{2}=0 \right ) = 1,\\
P\left ( y=1 | x_{1}=0, x_{2}=1 \right ) = 1, \nonumber\\
P\left ( y=0 | x_{1}=0, x_{2}=0 \right ) = 1.\nonumber
\end{align}

To show the recurrent property when samples following Eq.~(\ref{ANDlogicandxor2}) is given, we need to show that the absorbing states for Eq.~(\ref{ANDlogic}) disappears when ($x_{1}=1, x_{2}=0, y=1$) is given in addition, and the same applies for Eq.~(\ref{orlogic2}) when ($x_{1}=1, x_{2}=1, y=1$) is given. 

We first show that the absorbing state for ($x_1=1, x_2=1, y=1$) Eq.~(\ref{ANDlogic}), i.e., $\mathrm{TA}_{1}=\text{I}$, $\mathrm{TA}_{2}=\text{E}$, $\mathrm{TA}_{3}=\text{I}$, $\mathrm{TA}_{4}=\text{E}$, disappears when sub-pattern ($x_{1}=1, x_{2}=0, y=1$) is given in addition. Indeed, $\mathrm{TA}_{3}$ will move toward E when ($x_{1}=1, x_{2}=0, y=1$) is given, because

\hspace{0.55cm}\begin{minipage}{0.24\textwidth}
Condition:
$x_{1}=1$, $x_{2}=0$, $y=1$,  $\mathrm{TA}_{4}=\text{E}$.\\
Thus, Type I, $ x_{2}=0$, \\$C=x_{1} \wedge x_{2}=0$.
\end{minipage}
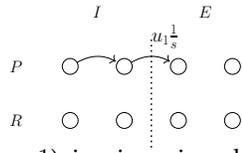
\resizebox{0.18\textwidth}{!}{
\begin{minipage}{0.24\textwidth}
\begin{tikzpicture}[node distance = .35cm, font=\Huge]
    \tikzstyle{every node}=[scale=0.35]
    
    \node[state] (E) at (1,1) {};
    \node[state] (F) at (2,1) {};
    \node[state] (G) at (3,1) {};
    \node[state] (H) at (4,1) {};
  
    \node[state] (A) at (1,2) {};
    \node[state] (B) at (2,2) {};
    \node[state] (C) at (3,2) {};
    \node[state] (D) at (4,2) {};
    
    \node[thick] at (0,1) {$R$};
    \node[thick] at (0,2) {$P$};
    \node[thick] at (1.5,3) {$I$};
    \node[thick] at (3.5,3) {$E$};
    
    \draw[dotted, thick] (2.5,0.5) -- (2.5,2.5);
    
    \draw[every loop]
    (A) edge[bend left] node [scale=1.2, above=0.1 of C] {} (B)
    (B) edge[bend left] node [scale=1.2, above=0.1 of C] {$~~~~~u_1\frac{1}{s}$} (C);

\end{tikzpicture}
\end{minipage}
}
Clearly, when ($x_{1}=1, x_{2}=0, y=1$) is given in addition,
$\mathrm{TA}_{3}$ has a non-zero probability to move towards “Exclude”. Therefore, “Include” is not the only direction that $\mathrm{TA}_{3}$ 
moves to upon the new input. In other words, ($x_{1}=1, x_{2}=0, y=1$) will make the state $\mathrm{TA}_{1}=\text{I}$, $\mathrm{TA}_{2}=\text{E}$, $\mathrm{TA}_{3}=\text{I}$, $\mathrm{TA}_{4}=\text{E}$, not
absorbing any longer. For other states, the newly added
training sample will not remove any transition from the
previous case. Therefore, the system will not have any new
absorbing state.

Following the same concept, we show that the absorbing state for ($ x_{1}=1, x_{2}=0, y=1$) shown in Eq.~(\ref{orlogic2}), i.e., $\mathrm{TA}_{1}=\text{I}$, $\mathrm{TA}_{2}=\text{E}$, $\mathrm{TA}_{3}=\text{E}$, $\mathrm{TA}_{4}=\text{I}$, disappears when sub-pattern ($x_{1}=1, x_{2}=1, y=1$) is given in addition. Indeed, $\mathrm{TA}_{4}$ will also move towards E when ($x_{1}=1, x_{2}=1, y=1$) is given, as:

\hspace{0.55cm}\begin{minipage}{0.24\textwidth}
Condition:
$x_{1}=1$, $x_{2}=1$, $y=1$,  $\mathrm{TA}_{3}=\text{E}$.\\
Thus, Type I, $ \neg x_{2}=0$, \\$C=x_{1} \wedge \neg x_{2}=0$.
\end{minipage}
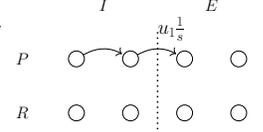
\resizebox{0.18\textwidth}{!}{
\begin{minipage}{0.24\textwidth}
\begin{tikzpicture}[node distance = .35cm, font=\Huge]
    \tikzstyle{every node}=[scale=0.35]
    
    \node[state] (E) at (1,1) {};
    \node[state] (F) at (2,1) {};
    \node[state] (G) at (3,1) {};
    \node[state] (H) at (4,1) {};
  
    \node[state] (A) at (1,2) {};
    \node[state] (B) at (2,2) {};
    \node[state] (C) at (3,2) {};
    \node[state] (D) at (4,2) {};
    
    \node[thick] at (0,1) {$R$};
    \node[thick] at (0,2) {$P$};
    \node[thick] at (1.5,3) {$I$};
    \node[thick] at (3.5,3) {$E$};
    
    \draw[dotted, thick] (2.5,0.5) -- (2.5,2.5);
    
    \draw[every loop]
    (A) edge[bend left] node [scale=1.2, above=0.1 of C] {} (B)
    (B) edge[bend left] node [scale=1.2, above=0.1 of C] {$~~~~~u_1\frac{1}{s}$} (C);

\end{tikzpicture}
\end{minipage}
}

Understandably, because of the newly added sub-patterns, the absorbing states in Eqs.~(\ref{ANDlogic}) and (\ref{orlogic2}) disappear and no new absorbing states are generated. In other words, the TM trained based on Eq.~(\ref{ANDlogicandxor2}), becomes recurrent.     

Following the same concept, we can show that the system becomes recurrent for Eqs.~(\ref{orlogic}), (\ref{ANDlogicandxor1}).
and (\ref{xor12}) as well.  For sake of conciseness, we will not provide the details here. 
  In general, any newly added sub-pattern will involve a probability for the learnt sub-pattern to move outside the learnt state and become recurrent.

\subsection{The Functionality of $T$ }


To encourage the TM to learn different sub-patterns and become possibly absorbed in the OR operator, parameter $T$ plays an important role. The functionality of $T$ is that it can block the training samples for learnt sub-pattern so that the clauses can be guided to learn un-learnt sub-patterns. 

\begin{mytheorem} \label{full6} The clauses can almost surely learn the OR logic in infinite time, when $ T\leq \floor{\frac{m}{2}}$. 
\end{mytheorem}

\textbf{Proof:}
To prove this theorem, we show that  (1) $T\leq \floor{m/2}$ is required so that the sum of the outputs of clauses for each sub-pattern can reach $T$.  (2) The system is absorbed only when the sum of the outputs of clauses for all sub-pattern reaches $T$ at the same time, i.e., $f_{\Sigma}(C_i(\bold{X}))=T$, $\forall \bold {X}= [x_1=0, x_2=1]$ or $[x_1=0, x_2=1]$ or $[x_1=0, x_2=1]$. (3) the absorbed point can follow all sub-patterns in OR logic. And (4) the input sample ($x_1=0, x_2=0$) will not have any possibility to give the sum greater than or equal to $T$. 

To prove (1), let us look at how to configure $T$ so that the sum of the outputs of clauses for each sub-pattern can reach $T$. The nature of the OR operator offers the possibility to represent 2 sub-patterns jointly. For example, $T$ clauses in the form of $x_1$ will result in the sum of the outputs as $T$ for both $(x_1=1, x_2=0)$ and $(x_1=1, x_2=1)$. If there are other $T$ clauses to represent the remaining sub-pattern,  2$T$ clauses can offer the sum of outputs as $T$ for any sub-pattern. Clearly, given $m$ clauses in total, when $T= \floor{m/2}$, all sub-patterns can possibly be covered.  When we have a smaller $T$, different sub-patterns may be represented by distinct clauses, offering more flexibility.  However, when $T> \floor{m/2}$, there will always be one or two sub-patterns that cannot obtain a sum of $T$ clauses. For this reason, the maximum $T$ value is $T= \floor{m/2}$. 

To prove (2), we show that the system is not absorbed when 0, 1 or 2 sub-patterns are blocked by $T$, and only when 3 sub-patterns are blocked by $T$, the system becomes absorbed. Clearly, when no sub-pattern is blocked, the training samples given to the system follow Eq.~(\ref{orlogic}). Following this type of training samples, 
it has already been shown in the previous subsection that the current system status is not absorbing. 

Now let's look at the case when only 1 sub-pattern is blocked. Clearly, if any $T$ clauses block only one sub-pattern, the system is updated based on Eqs.~(\ref{ANDlogicandxor2}),  (\ref{ANDlogicandxor1}), or (\ref{xor12}). As stated in the previous subsection, the current system status is not absorbing. 

We look at the cases when two sub-patterns are blocked but the third is not blocked. In other words, the sum of the outputs of clauses for any two sub-patterns reaches at least $T$, and the sum for the remaining sub-pattern is less than $T$.   In this case, only one type of the samples from Eqs.~(\ref{ANDlogic}) or (\ref{orlogic1}) or (\ref{orlogic2}) will be given to the TM. Based on the previous analysis of the equations, we understand that all clauses, including the ones that follow the two blocked sub-patterns, will be reinforced to learn the unblocked sub-pattern. This is due to the fact that only the samples following the unblocked sub-pattern are given to the system, and in this circumstance, the system has only one absorbing state, which follows the given training sample.   
This property will encourage the clauses that have learnt the blocked sub-patterns move out of the current states. Once they are out of their current states before the sum of the third one also reaches $T$, the blocked sub-patterns will be unblocked and the system becomes one of three cases described by Eqs.~(\ref{ANDlogicandxor2}), (\ref{ANDlogicandxor1}) or (\ref{xor12}).  Clearly, the system status is still not absorbing. 

In fact, the system will only be absorbed when all three sub-patterns are blocked at the same time. In this situation, no Type I feedback is given and only Type II feedback can possibly update the system. Type II feedback is only triggered by ($x_1=0$, $x_2=0$, $y=0$) in OR operator. For Type II feedback, based on Table~\ref{table:type_ii_feedback}, any transition is only triggered as a penalty when excluded literal has 0 value and the clause is evaluated as 1. Specifically for OR operation, this only happens when $C=\neg x_1\wedge \neg x_2$ or $C=\neg x_1$ or $C=\neg x_2$.  For $C=\neg x_1 \wedge \neg x_2$, based on the Type II feedback, the TA with the action ``excluding $x_1$" and the TA with the action ``excluding $x_2$" will be penalized. In other words, the actions of the TAs for $x_1$ and $x_2$ will be encouraged to move from exclude to include side. As soon as any one of TAs for $x_1$ or $x_2$ (or occasionally both of them) becomes included, the clause will become $C =  \neg x_1 \wedge x_1 \wedge \neg x_2$ or $C =  \neg x_1 \wedge \neg x_2 \wedge x_2$ (or occasionally $C =  \neg x_1 \wedge x_1 \wedge x_2 \wedge \neg x_2$). In this case, input ($x_1=0$, $x_2=0$) will always result in 0 as the output of the clause and then the Type II feedback will not update the system any longer.  Following the same concept, for $C=\neg x_2$, the Type II feedback will encourage the excluded $x_1$ to be included so that the clause becomes $C=x_1\wedge \neg x_2$. The same applies to $C=\neg x_1$, which will eventually become $C=\neg x_1\wedge x_2$ upon Type II feedback. When all clauses in $C=\neg x_2$ or $C=\neg x_1$ are also updated to $C=x_1\wedge \neg x_2$ or $C=\neg x_1\wedge  x_2$, the system is absorbed because no feedback is triggered up on any input sample.

We summarize the requirement for an absorbing state. 
\begin{itemize}
    \item For any sample $\bold{X}$ that satisfies OR, i.e. $\bold{X}=[x_1=1, x_2=1]$, or $\bold{X}=[x_1=1, x_2=0]$, or $\bold{X}=[x_1=0, x_2=1]$, the sum of the outputs of clauses, i.e., $f_{\Sigma}(C^i(\bold{X}))$ must be at least $T$. This will block any Type I feedback.
    \item There are no clauses with only negated literal, such as $C=\neg x_1$ or $C=\neg x_2$ or or $C=\neg x_1\wedge\neg x_2$. This will block any Type II feedback. 
\end{itemize}

(3) Now we prove that all the absorbed points can cover all three sub-patterns in OR logic. 

Here we employ proof by contradiction. Suppose there is an absorbing state that does not include all sub-patterns in the OR logic. Let's name the not-included sub-pattern\footnote{There might be more than 1 not-included sub-patterns, but the analysis concept is the same.} as sub-pattern A. For this reason,  for any input samples belong to sub-pattern A, the clauses in the absorbing state will not output 1. Therefore, the sum of the clauses for sub-pattern A will not reach $T$, which is a conflict with the requirement for the absorbing state, where the TM needs a sum of at least $T$ clauses for any sub-pattern.     

(4) We now show that input sample $(x_1=0, x_2=0)$ will not give a sum of clause outputs greater than or equal to $T$. This is to avoid any possible false positive case upon input $(x_1=0, x_2=0)$ in the testing phase.  Obviously, to have a positive output, the clause should be in the form of $C=\neg x_1$ or $C=\neg x_2$ or $C=\neg x_1 \wedge \neg x_2$. It has already shown in (2) that Type II feedback will eliminate such clauses. For this reason, $(x_1=0, x_2=0)$ will never result in a sum of clause outputs greater than or equal to $T$.  

Before the system is absorbed, the system moves back and forth in the intermediate states. As long as $T\leq \floor{m/2}$ holds, the system will eventually be absorbed. 
We thus have the OR logic almost surely and conclude the proof. \hfill \qedsymbol


Now let us take a look at one example where $m=7$, $T=3$. Due to the randomness of the learning nature, the system may be absorbed in one of numerous absorbing states. Here we list 3 representative ones. 
\begin{itemize}
    \item $C_1=C_2=C_3=x_1$, $C_4=C_5=C_6=x_2$, $C_7=*$. 
    \item $C_1=C_2=C_3=x_1$, $C_4=C_5=C_6=\neg x_1\wedge x_2$, $C_7=*$.  
    \item $C_1=C_2=x_1$, $C_3=x_1\wedge \neg x_2$ $C_3=C_4=\neg x_1\wedge x_2$, $C_6=x_2$, $C_7=*$.  
\end{itemize}
Here $*$ means the clause can be in any form other than $\neg x_1$, $\neg x_2$, or $\neg x_1 \wedge \neg x_2$. Note that all above examples can sum up to at least $T$ for any input $\bold{X}$ in OR logic.  
\begin{remark}
\label{remark2}
When $T$ is greater than half of the number of the clauses, i.e., $T>\floor{m/2}$, the system will not have an absorbing state. We conjuncture that the system can still learn the sub-patterns in a balanced manner, as long as $T$ is not configured too close to the total number of clauses $m$,  $s$ is large, and the  number of states in the TAs is also large. 
\end{remark}

Given $T>\floor{m/2}$ the system will oscillate. Nevertheless, with high probability, the system will have at least $m-T$ clauses that follow each sub-pattern, especially when $s$ is large. First of all, when $s$ is large, the clause is less likely to get out of the learnt sub-pattern due to a training sample from the conflicting sub-pattern. This is because of the nature of TM. In more detail, the probability for an included literal in a clause that has learnt a certain sub-pattern to change towards the other sub-pattern is $u_1/s$, which only happens when a training sample of the other sub-pattern is given. On the contrary, the reward is $u_1\frac{s-1}{s}$ if a training sample of the same sub-pattern is received. $\frac{s-1}{s}>\frac{1}{s}$ holds if $s$ is large. 
Similarly, when the number of TA  states is large, the probability of being in the include side of the TAs is also large given $\frac{s-1}{s}>\frac{1}{s}$. 

\subsection{Revisit the XOR Operator}

Let us revisit the proof of XOR operator. As stated in~\cite{jiao2021convergence}, when the system is absorbed, the clauses follow the format $C=x_1\wedge \neg x_2$ or $C=\neg x_1\wedge x_2$ precisely. In other words, a clause with just one literal, such as $C=x_1$, cannot absorb the system. The main reason is that the sub-patterns in XOR operator are mutual exclusive, i.e., the sub-patterns cannot be merged in any way. Although Type I feedback can be blocked when $T$ clauses follow one sub-pattern using one literal, the Type II feedback can reinforce the other missing literal to be included.
For example, when $T$ clauses happens to converge to $C=x_1$, the Type I feedback from any input samples of $(x_1=1, x_2=0, y=1)$ will be blocked. In this situation, the unblocked Type II feedback from $(x_1=1, x_2=1, y=0)$ will encourage the clause to include $\neg x_2$. This is because upon a sample $(x_1=1, x_2=1, y=0)$,  we have Type II feedback, $C=x_1=1$, and the studied literal is $\neg x_2=0$. When the TA for excluding $\neg x_2$ is considered, a big penalty is given to the TA, making it moving towards action Included, and thus $C=x_1$ eventually becomes $C=x_1\wedge \neg x_2$. Following the same concept, we can analyze the development for $C=\neg x_1$, $C=x_2$, and $C=\neg x_2$, which will eventually converge to $C=\neg x_1\wedge x_2$ or $C=x_1\wedge \neg x_2$, upon Type II feedback. 

\section{Conclusions}\label{conclusions}
In this article, we prove the convergence of the TM for the AND and the OR operators. Together with the proofs in~\cite{zhang2020convergence} and~\cite{jiao2021convergence}, we complete the convergence analyses of TM for the fundamental digital operators, which lay the foundation for future applications and analytical studies of TM. 

\section*{Acknowledgements}
This work is supported in part by the project Spacetime Vision: Towards Unsupervised Learning in the 4D World financed by the EEA and Norway Grants 2014-2021 under the grant number EEA-RO-NO-2018-04. This work was also
supported by the Research Council of Norway through
the AI4Citizen project (320783) and the AIEverywhere
project~(312434).
\bibliographystyle{IEEEtran}
\bibliography{main}
\end{document}